# Pattern-wave model of brain function. Mechanisms of information processing.

Author: Alexey Redozubov

The structure of the axon-dendrite connections of neurons of the brain creates a rich spatial structure in which various combinations of signals surrounding neurons are present. Structure of dendritic trees and shape of dendritic spines allow to repeatedly increase combinatorial component through cross synapses influence neighboring neurons. In this paper it is shown that the diffuse spreading of neurotransmitters allows neurons to detect and remember significant set of environmental activity patterns. Extrasynaptic metabotropic receptors clusters are the core element of that. The described mechanism leads to the appearance of the wave processes, based on the propagation of the front-line areas of spontaneous activity. In the proposed model, any compact pattern of neural activity is seen as a source emitting a diverging wave endogenous spikes. It is shown that the spike pattern of the wave front is strictly unique and has uniquely defined pattern that started the wave. The propagation of the waves with a unique pattern allows anywhere in brain by passing by wave patterns judge the whole brain information processes. In these assumptions naturally described mechanism of projection information between regions of the cortex. Performed computer simulations show the higher information effectiveness of such model.

## Overview

Brain handles information. Perhaps, it is the only thing that can be said about the brain with some confidence. In what form this information is represented and how operations occur on it - today, it is the open question.

Similar to the brain, computers are also able to process information. It is clear how information comes into a computer, what forms information takes, how it memorized and based on what principles converted. Of course, it is very tempting to use the computer analogy for the brain design and try to find there something similar to the computer modules and programs. But, unfortunately, the architecture of the brain is so different from the computer architecture that a direct comparison doesn’t add understanding how the brain works.

When there’s no satisfactory explanatory theory available, we have to operate the ideas which, although do not offer a complete answer, but indicate the directions of searches that may possibly lead to success. In traditional neuroscience there are two main concepts that claim to explain how the brain processes information.

The first concept is the idea of the "grandmother neurons" and patterns of neural activity. This concept is well illustrated by the many artificial neural network (eg multilayer perceptron (Rosenblatt F., 1962) or convolutional network (Y. LeCun, Y. Bengio, 1995)). In models where neurons’ synaptic weights adjusting, neuron acquire the ability to detect certain statistical properties of the input information. Depending on the type of information that is supplied to the neural network layer, the neurons during learning process acquire a certain specialization. As a result,  such neurons can be associate with certain concepts, which they are detecting. Set of neurons to respond to the current image, creates a pattern of activity, which describes input information.

Then these ideas applied to a real brain design: layers of artificial neural network are mapped to cortical regions, and the activity of formal neurons is compared with evoked neuronal activity. Such comparison highlights few problems.

The contradiction in the fact that the pattern coding required transferring of the whole picture of activity from one zone to another, but projection fiber bundles of real brains clearly do not meet this requirement by projection volume and connection properties (Pitts W., McCulloch WS, 1947).

Information integration on more generalized cortex areas requires enlarge of neurons tracking field, which is not observed in the actual cortex. The needs to track not just local receptive field, by all prior cortex led to models like neocognitron Fukushima (Fukushima, 1980). In this model the plane of simple cells used to monitor invariant patterns over the entire space of the cortex. However, this approach is valid weak biologically. It is applicable in fact, only for simulation of the visual cortex, where thus attempt association image invariants.

The second concept - spike networks (eg, Izhikevich, 2007). Since the activity of neurons is series of impulses, it is assumed that the information is encoded by time intervals between the spikes and their pattern. In this case, the space of cortex is perceived as a medium, which allows to distribute impulse sequences, while retaining their basic temporal characteristics.

This correlates well with the fact that brain activity is accompanied by a rhythmic activity detected on EEG. Also shown, the wave propagation of activity in cortex. (Michael T. Lippert, Kentaroh Takagaki, Weifeng Xu, Xiaoying Huang, Jian-Young Wu, 2007). In addition, there are enough weighty arguments in favor of the fact that the basis of information processes in the brain and the event memory have holographic principles (Pribram, 1971). Spike model, since it allows the description of interference phenomena, gives some hope for progress in this direction.

However, the spike networks have not been able to show quite strong and noiseproof mechanisms of information processing. Furthermore, spiking network suggest an analog description of information instead of the information description in discrete terms. Discrete description is one of the key moments of the cognitive approach. Abandoning it highly undesirable.

It should be noted that none of these approaches has not yet allowed us to construct a convincing model of associative memory. It is tempting to assume that the properties of neurons to detect certain images and memories of specific events coded by change in synaptic weights of real neurons. Hebbian learning (Hebb, 1949) and many other methods used for training neural networks, is really possible to obtain interesting results, but they come into a conflict with the requirements for the safety of the old experience. Stability plasticity dilemma is solved well in adaptive resonance theory Grossberg (Grossberg, 1987), but the traditional neural network adaptive resonance have a fairly weak biological plausibility. The result is a situation where attempts to build biologically accurate model only within the synaptic plasticity is not yet give satisfactory result. Involvement additional mechanisms require, on the one hand, biological justification, and on the other hand, it makes sense only if allows us to give an entirely new interpretation of information processes.

In this paper we will describe the principle of building a biologically accurate neural networks using except evoked activity also endogenous (spontaneous) activity mechanisms. Computer modeling of these networks showed their high information capacity. It turned out that they quite naturally realized many information mechanisms inherent in the real brain, including: generalization, construction of distributed associative memory and reinforcement learning procedure.

## Evoked activity

In the resting state between the internal and external environment of the neuron potential difference (membrane potential) is about 70 millivolts. It is formed by the protein molecules that act as ion pumps. As a result, the membrane acquires a polarization at which the negative charge is accumulated inside the cells, and positive outside.

Neuron surface covered branching outgrowths - dendrites. To the body of the neuron and it dendrites adjoin axon’s closure of other neurons. Their joints called synapses. Through synaptic interaction neuron is able to respond to incoming signals and in certain circumstances to generate its own impulse, called a spike.

Signal transmission at synapses occurs due to release of neurotransmitters. When a nerve impulse along the axon enters the presynaptic terminal, it releases from the synaptic vesicle neurotransmitter molecules typical to that synapse. On the membrane of a neuron that receives the signal placed receptors that interact with that neurotransmitter.

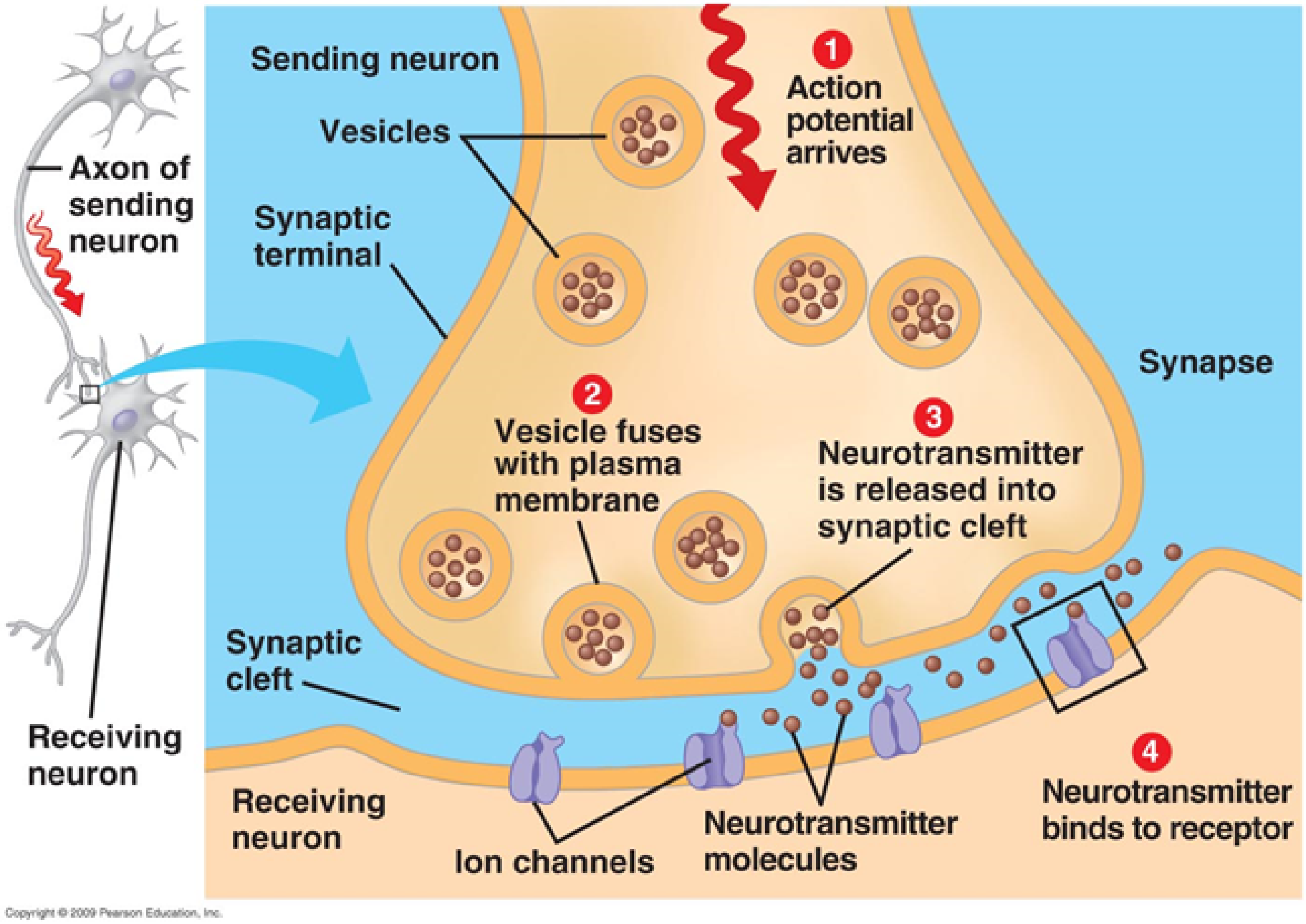

Figure 1. Chemical synapse

Receptors located in the synaptic cleft are mostly ionotropic. They are ion channels, able to move ions through the membrane of the neuron. Neurotransmitters act on the receptors so that their ion channels open. As a result, the membrane either depolarized or hyperpolarized - depending on which channels are affected, and accordingly, what type of this synapse. In excitatory synapses open channels, mostly permeable cations into the cell - the membrane is depolarized. In excitatory synapses open channels, mostly permeable cations into the cell - the membrane is depolarized. In braking synapses open channels that output the cations out of the cell, which leads to hyperpolarization of the membrane.

In certain circumstances, the synapses can change their sensitivity is called synaptic plasticity. This leads to the fact that some of the synapses become more and others less sensitive to external signals.

Simultaneously, the neuron's synapses receives many signals. Inhibitory synapses increase the membrane potential that lead to hyperpolarizing. Activating synapses, on the contrary, trying to defuse the neuron. When the total depolarization exceeds a threshold of initiation, discharge happens, called an action potential, or spike.

After the release of neurotransmitters special mechanisms ensure their recycling and re-uptake, which leads to clearing the synaptic cleft and the surrounding synapse space. During the refractory period following a spike, the neuron is not able to generate new impulses. This period determines the maximum frequency of spike generation for that neuron.

## Combinatorial structure of the dendrite

When an action potential propagating along the axon, reaches the recipient neuron, it causes the release of neurotransmitters into the synaptic cleft. These mediators determine the synapse contribution to the overall change of the membrane potential of a neuron that receives the signal. But the part of mediators falls outside the synaptic cleft and filling the space formed by other neurons and their surrounding glial cells. This phenomenon is called spillover (Kullmann, 2000). Moreover, mediators are emitted none synaptic axon terminals and glial cells (Figure 2). Concentrations of neurotransmitter outside of synaptic clefts is to be much less than in synaptic clefts. This concentration is not enough to significantly affect the state of the membrane potential through direct activation mechanisms. But in some cases it is enough to activate neuron through metabotropic receptors (including second messenger systems) cause endogenous (spontaneous) spikes.

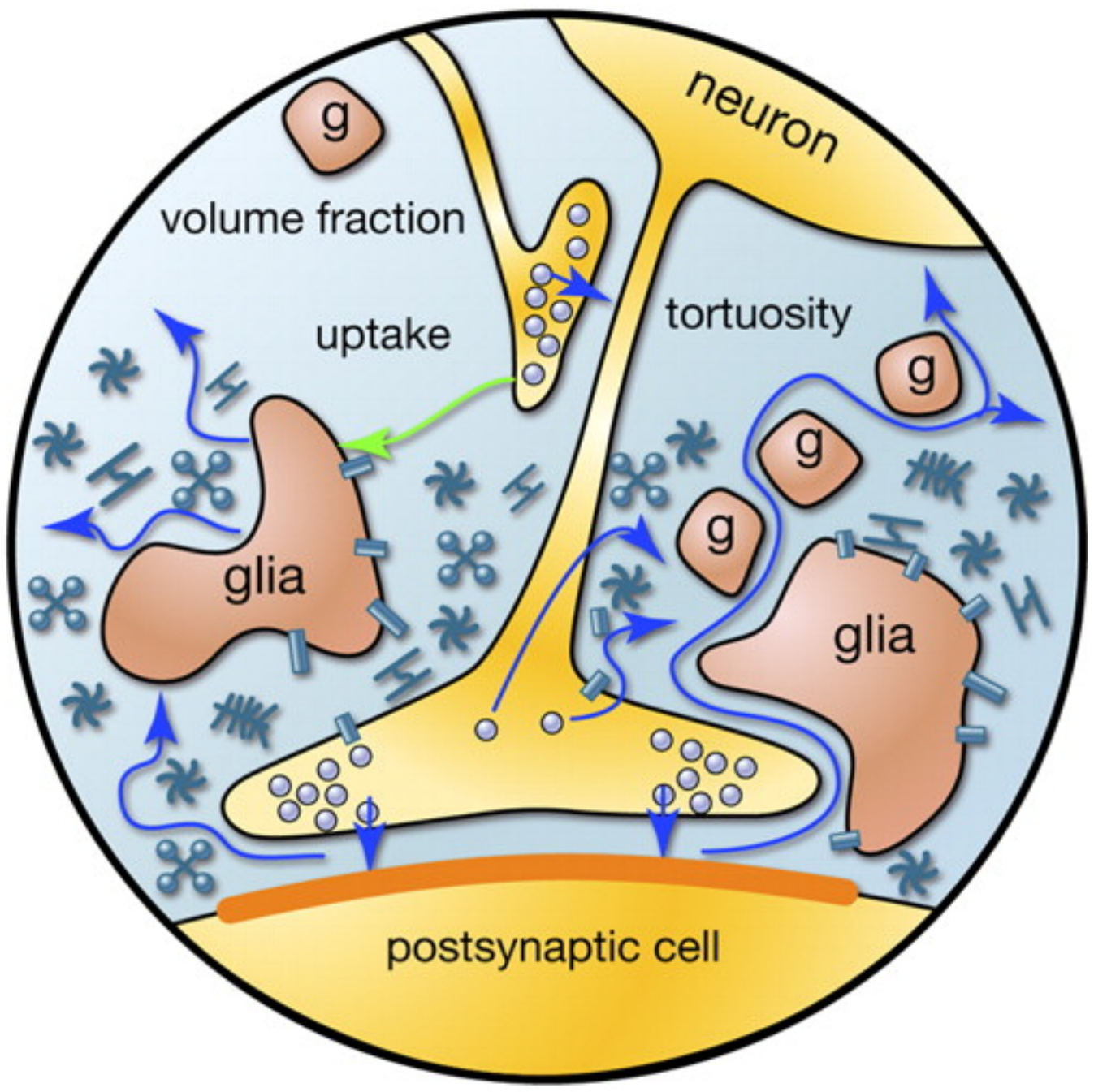

Figure 2. Sources neuromediators outside the synaptic cleft (Sykova E., Mazel T., Vagrova L., Vorisek I., Prokopova-Kubinova S., 2000)

Try to estimate the quantity and structure of neurotransmitters diffuse sources. For this we use quantitative estimates of the parameters of the cortex, in table:

Table 1. Summary of parameters obtained in the study of mouse brain (py-cell - pyramidal cell, Type-I - synapses between pyramidal cells) (Braitenberg V., Schuz A., 1998)

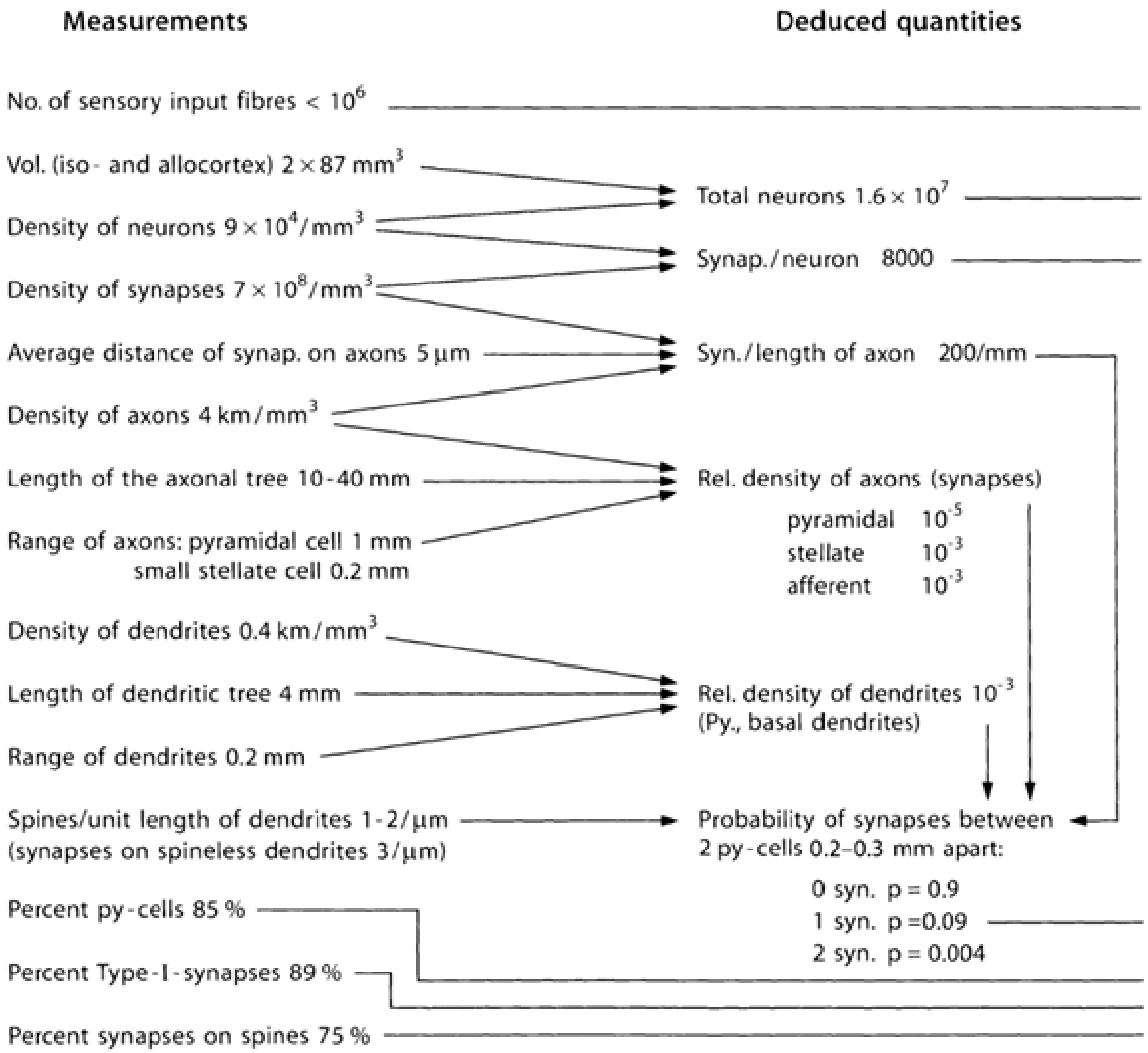

Most of the synapses (90-95%) are in the dendritic tree of the neuron. For different types of neurons form dendritic trees are different. Same time, the general principle is preserved: the dendritic tree consists of a set of branching outgrowth, with the highest density of synaptic connections of neuron in small spatial region. For basic types of neurons it is about 200 microns (Figure 3, Figure 4).

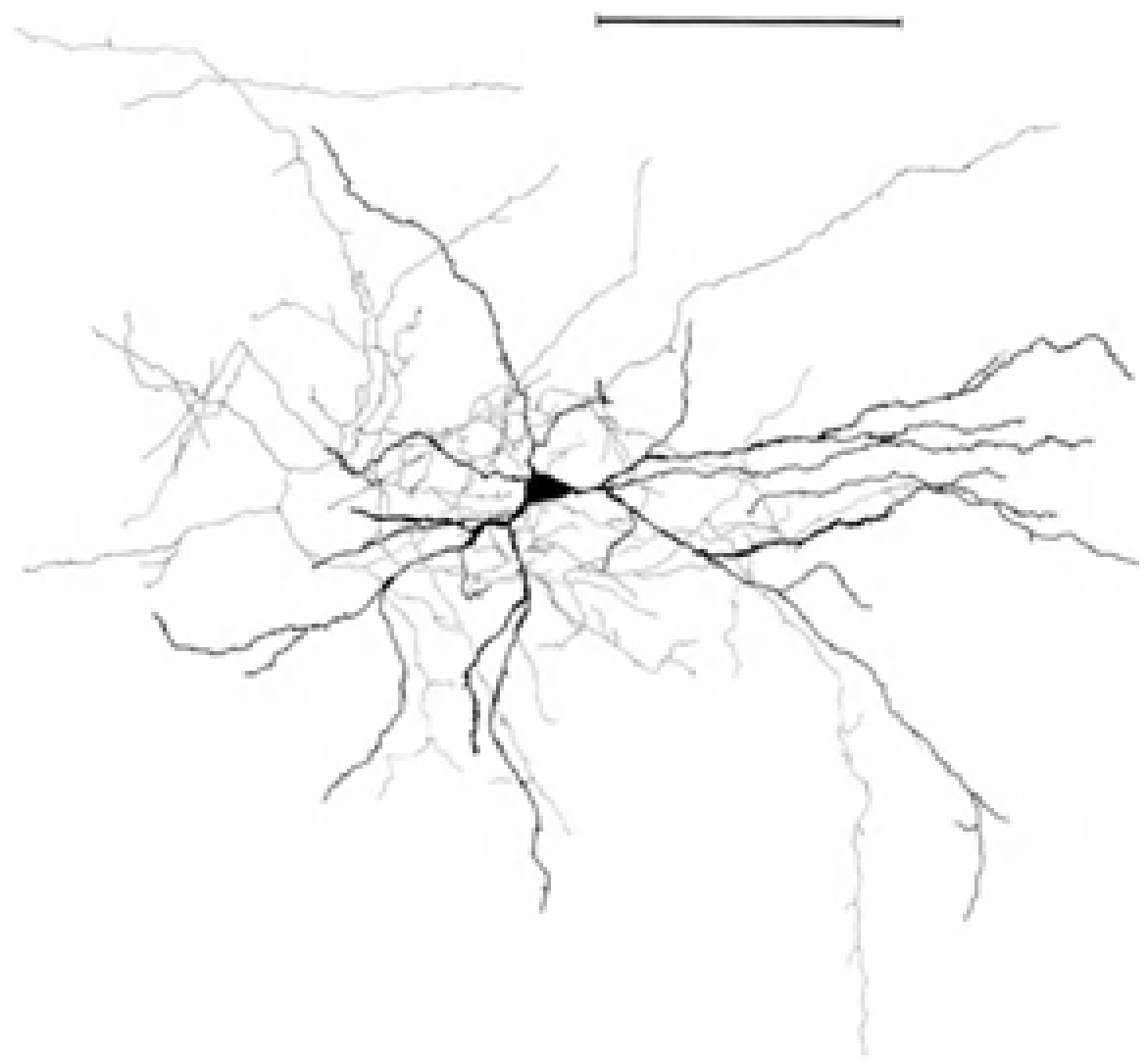

Figure 3. Stellate neuron structure, ruler - 0.1 mm (Braitenberg, 1978)

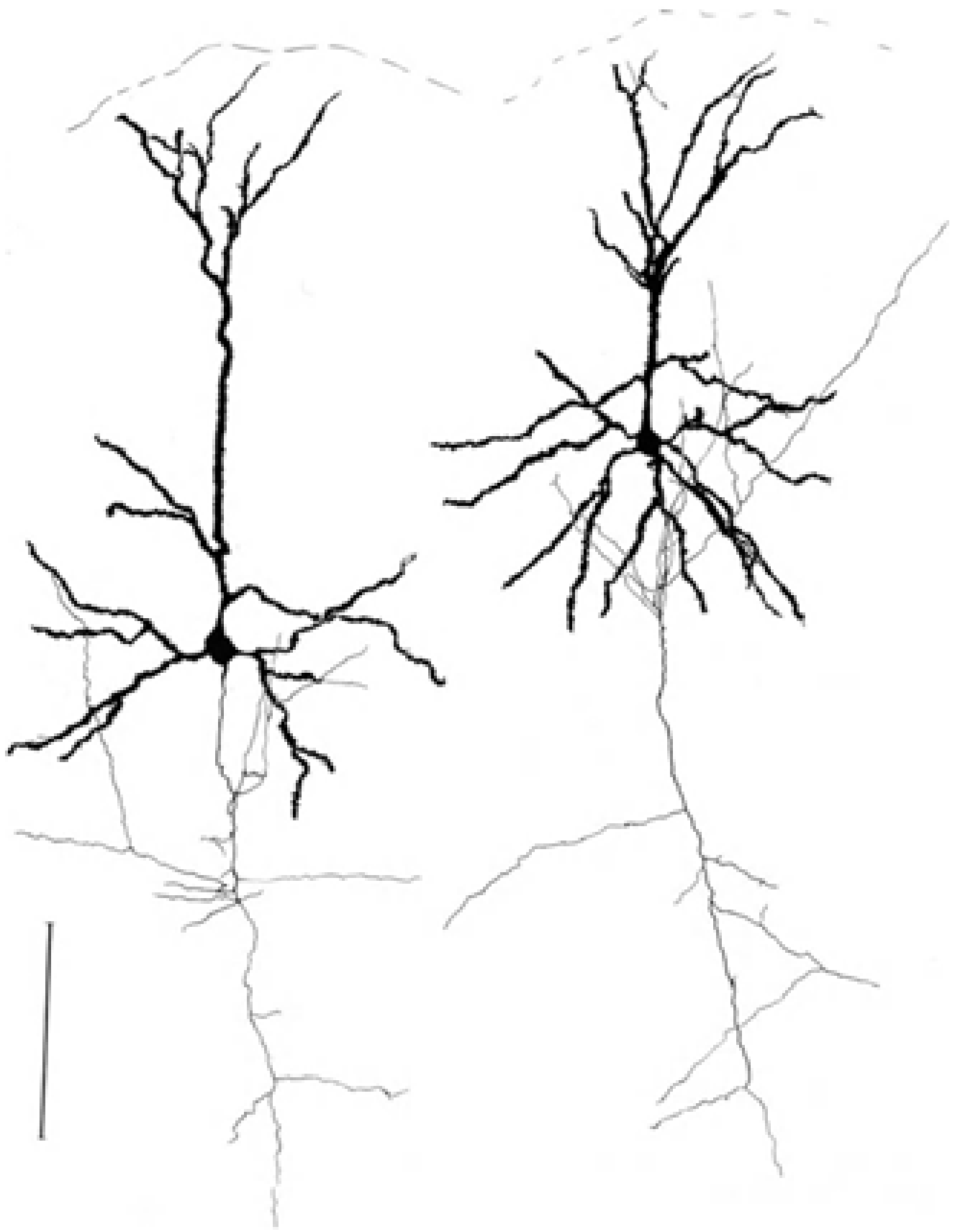

Figure 4. Structure of pyramidal neurons, ruler - 0.1 mm (Braitenberg, 1978)

Most of synapses (75%) located on the dendritic spines. That is typical for pyramidal cells (Figure 5). The total length of the dendritic tree branches of a single neuron - 4 mm. The average distance between synapses on dendritic branch 0.5 microns.

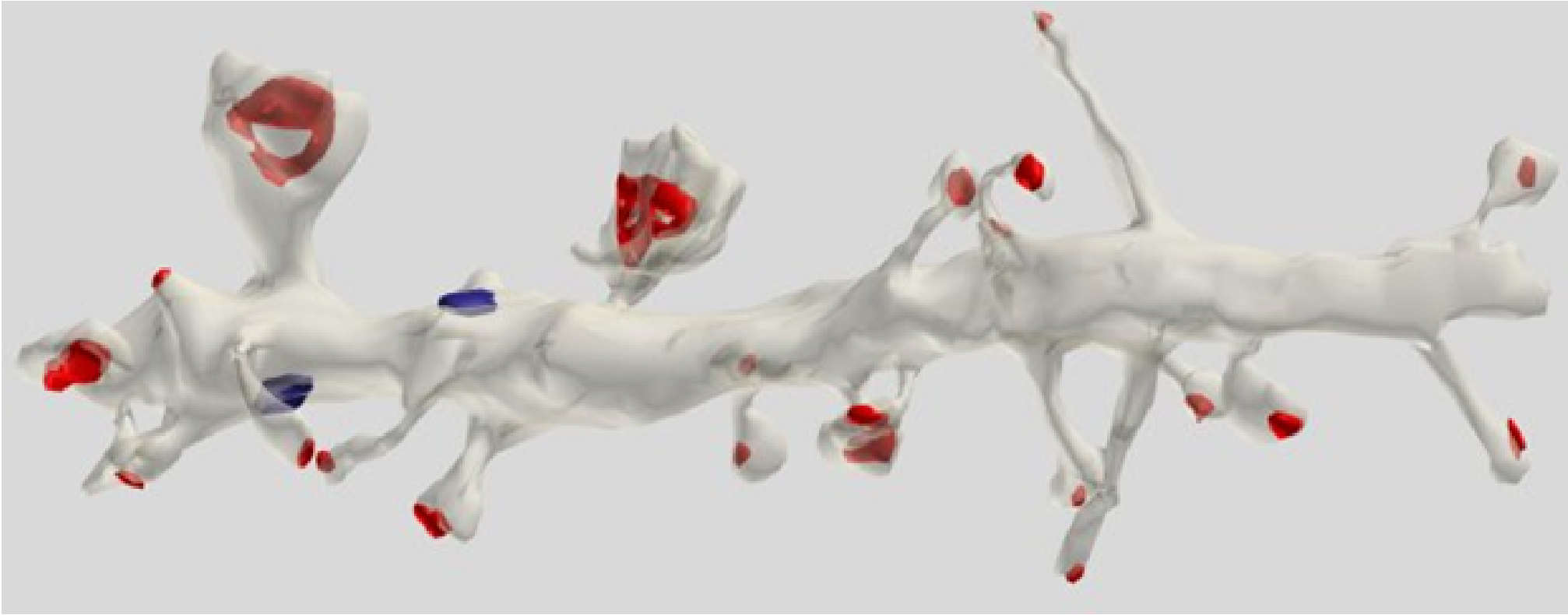

Figure 5. Segment dendrites of pyramidal cells. Red marked synapses on spines, blue - on the dendritic trunk (Dr. Kristen M. Harris)

Computer simulations performed on the basis of real anatomical and physiological data indicated that, for example, glutamate may extend beyond the synaptic cleft in amounts sufficient to activate the NMDA receptor in a radius commensurate with the distance between neighboring synapses (0.5 microns) (Rusakov DA, Kullmann DM, 1998). It can be assumed that a significant concentration of neurotransmitters after spillover observed at the site of the dendrite length of about 1-2 microns. In such a region may be located from two to four synapses belonging to the dendrite.

If you take a portion of the dendrite length 5 um (Figure 6), the expected number of synapses on it will be about 10.

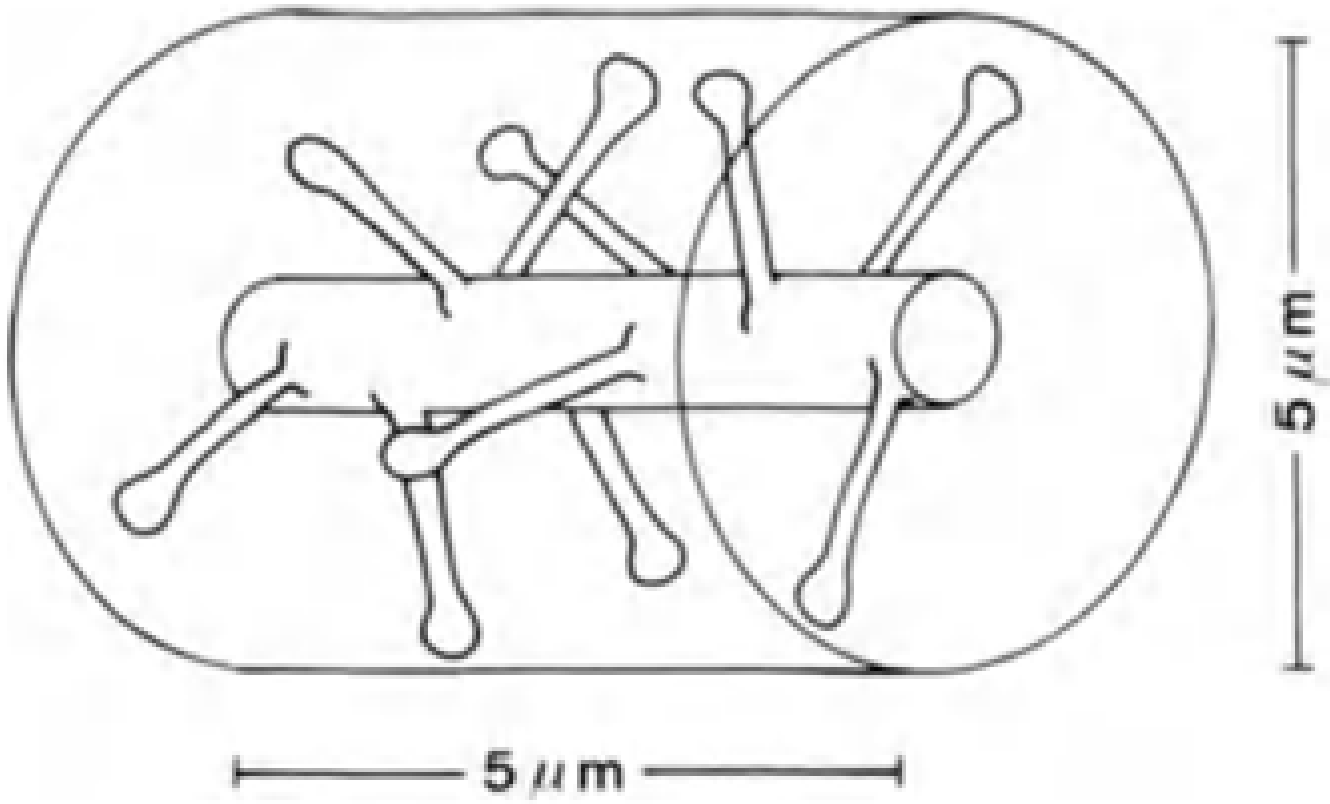

Figure 6. Portion of dendrite (Braitenberg V., Schuz A., 1998)

But dendritic branches of neurons are closely intertwined with branches of other neurons. They are very close to each other. Due to the height of dendritic spines, synapses of one dendrite, may be closer to the surface of another dendrite than it own synapses.

If synapses were uniformly distributed in the space of cortex, then volume of cylinder with 5 um height and 5 um diameter (pic 6) and on density distribution of synapses $7x10^8$/$мм^3$ would have approximately 100 synapses. That is 10 times more than synapses of same dendrite. In fact, a substantial part of brain volume occupied by glial cells and neurons body, that further increases the packing density of synapses.

We consider the number of synapses neighboring the concrete place of the dendritic surface, to assessing how many sources can affect the extrasynaptic density of neurotransmitter. Here, apart from the above, must remember that, some synapses can emit neuromediators, other than those which are sensitive metabotropic receptors on the dendrite of interest to us. And secondly, that extrasynaptic axon terminals are additional source of neurotransmitter. If the first lowers the amount of neurotransmitter sources, the second raises it.

It is difficult to precisely quantify the density of sources. For a rough estimate let's assume that the total density of sources five times more than the density of synapses on dendrite of single neuron. So, the number of sources, that significantly affects the formation of neurotransmitters density in each place of the dendrite, somewhere around 10-20 elements. Ie, the activity of 10-20 surrounding neurons or collaterals afferent axons forms density neuromediators in each individual place of dendritic tree.

The average distance between synapses for axon is 5 micrometers. It's several times larger than mediators propagation distance beyond synapses. This greatly decreases the probability that two synapse of neuron are simultaneous affect same volume of dendrite surface.

Two pyramidal neurons located at a distance of 0.2-0.3 mm have one common synapse with estimated probability only 10%. At the same time on pyramidal neurons accounted for 85% of all synapses in the cortex. Respectively, 75% of synaptic contacts - is contact between the pyramidal cells (Braitenberg V., Schuz A., 1998). This means that the majority of synaptic contacts of neurons falls on their immediate environment, ie neurons on distance no more than 100-150 um along the plane of the cortex. The thickness of cortex due to the vertical orientation of the apical dendrites of pyramidal cells, this distance will be slightly more. Moreover, the specificity of neighboring neurons connections allow two neurons in the immediate vicinity connect by set of synaptic contacts.

A rough estimate of the number of neurons having multiple communication with a selected neuron can be obtained by the number of neurons falling in a sphere of radius about 120 um. The volume of this sphere is 0.007 $mm^3$. At density of neurons $9 \cdot 10^4/mm^3$ we get 650 neurons. Very rough estimate of the number of synapses relating to these neurons can be obtained if we take the number of synapses per neuron, falling in that volume. If present dendritic tree as aggregate of branches emanating radially from a common center, and length R. Distribution of synapses is uniform along the length of the branches. Then volume with radius r have total number of synapses equal r / R. f neuron in average have 8,000 synapses and the span of dendrites is 0.2 mm, there be 5,000 synapses.

It's possible to say that on average nearby neurons have eight synaptic contacts with each other. This agrees fairly well with the estimate 5.6, obtained by direct counting of contacts between neurons layer V (13 neurons pairs were tracked) using an optical microscope (Liibke J., Markram H., Frotscher M., Sakmann B., 1996).

Now let's try to understand the meaning of such a structure relations from the viewpoint of the density distribution of extrasynaptic mediator. For this lets use a simplified model. Let's take neuron and enumerate neurons in volume surrounding selected neuron. Each of these enumerated neurons will have:

- a several synaptic contacts with the dendrites of selected neuron;
- a few places where it synapses connected to other neurons are located in the immediate vicinity of the selected neuron dendrite;
- a number of extrasynaptic axon's terminals adjacent to the selected neuronal dendrite (mainly true for candelabrum cells).

For simplicity, all sources of neurotransmitters are equal. Total number of sources can be estimated using the previously adopted coefficient. Again, for simplicity, the unequal distribution of connections in space will not take into account.

It possible to represent dendritic tree as a long branch uniformly distributed sources (figure 7). For each source on that branch, the number of the neuron from the surrounding area, responsible for it, can specified. Each of the neurons around will have several contacts randomly distributed along the dendrites. Denote this mapping by vector D with elements $d_i$.

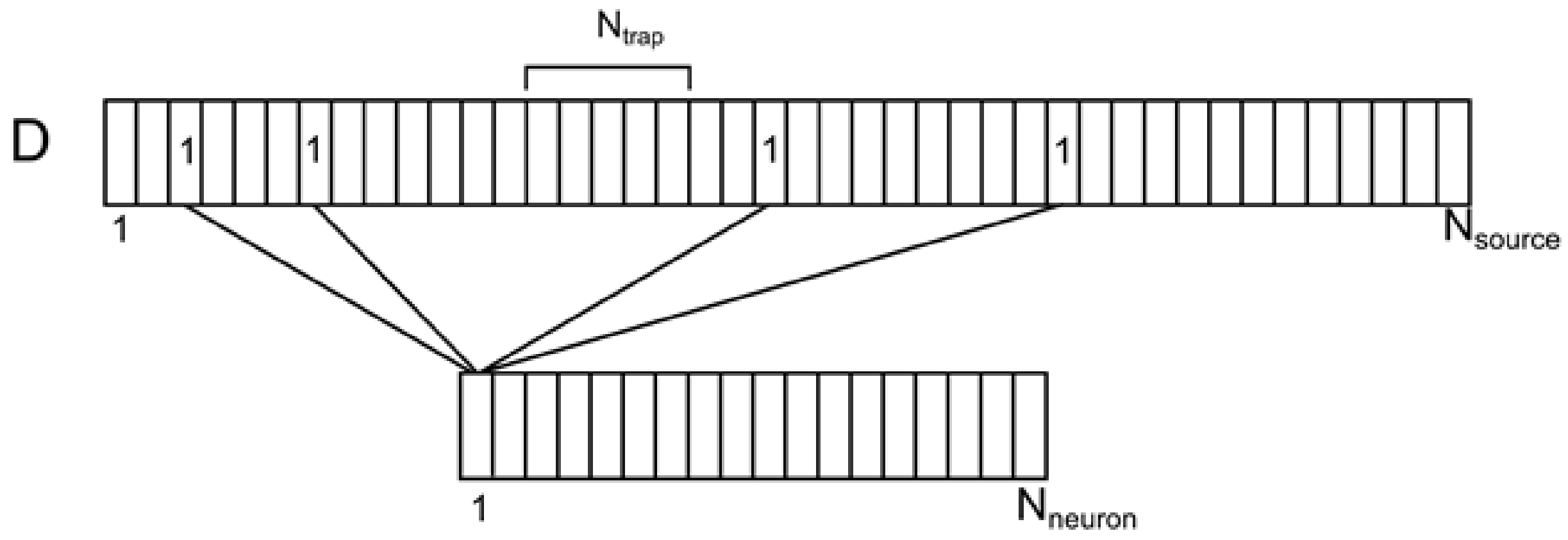

Figure 7. Correspondence surrounding neurons and their dendrite contacts

Where:
$N_{neuron}$ – number of neurons around selected one
$N_{source}$ – number of sources of single neuron
$N_{trap}$ – number of sources, that affect density level of neurotransmitters (synaptic trap)

Now suppose that a few neurons around selected one generate spikes. This can be taken as a signal that selected neuron see. Denote $N_{sig}$ - the number of active neurons, creating an information signal. Binary vector S is that signal.

For all positions in the dendrite except border one, the mediator density can be calculated by the formula

$$\boldsymbol{p}_i = \sum_{k=i-\left[\frac{N_{trap}}{2}\right]}^{i-\left[\frac{N_{trap}}{2}\right]+N_{trap}-1} S_{d_k}$$

For example, for the signal shown in the figure 8, marked synaptic density in the trap is 2 (the sum of the signals from the 1st and 4th neurons).

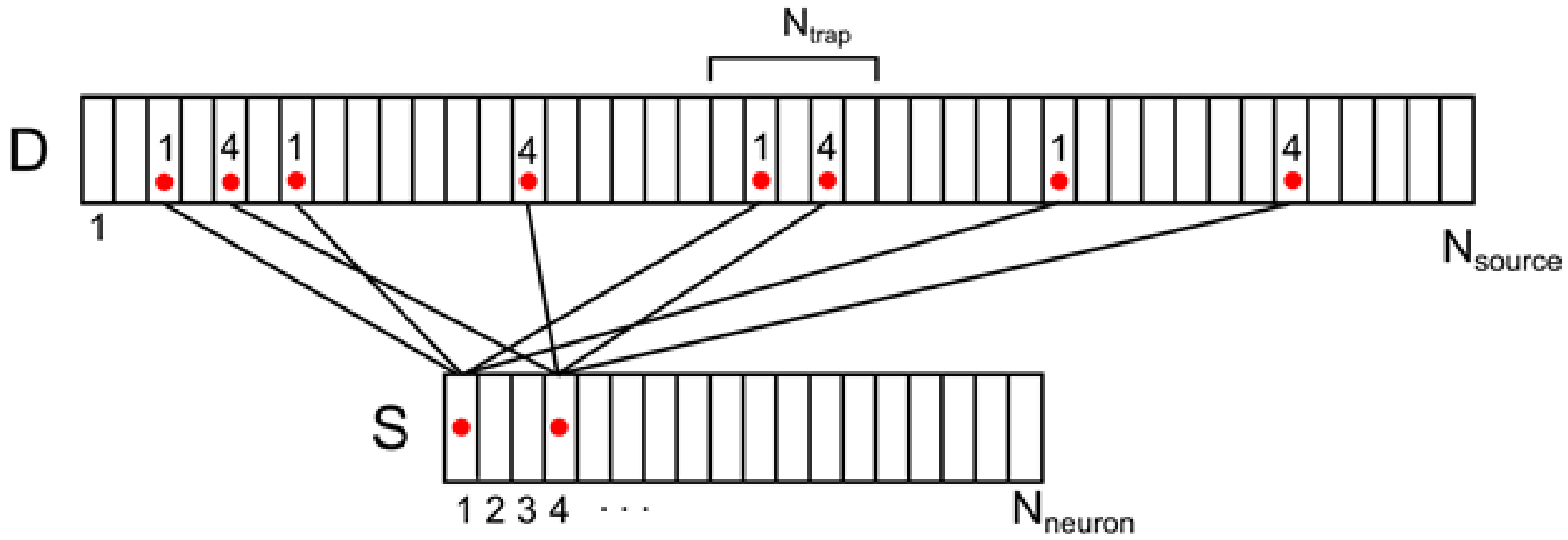

Figure 8. Showing Activity of two neurons surroundings on the dendritic tree (only a portion of bonds and numbering)

For any arbitrary signal can calculate the distribution pattern of mediators density on the dendrite. This density range between $0$ and $N_{trap}$. The maximum value will be achieved if all sources, forming synaptic trap, were active.

Let's use the averaged values of parameters specific for real rat cortex (Braitenberg V., Schuz A., 1998). On the basis of these and earlier assumptions about the number of sources, the following model parameters obtained:
$N_{neuron}$ = 650
$N_{source}$ = 25000
$N_{trap}$ = 15
Assume that the signal is encoded by activity, for example, 1.5% of cortical neurons, then $N_{sig}$ = 10

It is easy to calculate the probability to find the place on dendrite where density of mediator exactly K for arbitrary signal of $N_{sig}$ active neurons. For the above parameters the probability takes the following values:

Table 2. Table probability of finding at least one trap with a given density.
The first column - the desired amount of the active sources in the trap. Second - the probability of finding at least one place on dendrite where given number of active sources

| **K** | **P** |
|---|---|
| 0 | 0.984 |
| 1 | 1.000 |
| 2 | 1.000 |
| 3 | 0.996 |
| 4 | 0.287 |
| 5 | 0.016 |
| 6 | 0.001 |
| 7 | 0.000 |
| 8 | 0.000 |
| 9 | 0.000 |
| 10 | 0.000 |
| … | |

When parameters are close to the actual configuration of the cortex, for any signal, created by about 1.5% of the neurons, the following is true:

- There is about 1.6% of the neurons, which have trap on the dendrites, where 50% of axons' signals intersects;

- Almost every neuron have trap, where at least 30% axons' signals intersects.

The meaning of this result is very interesting. Suppose, that the information in the cortex somehow encoded simultaneous activity of a relatively small amount $(N_{sig})$ compactly located neurons. This is not the whole brain activity, but information processes in a small volume where neurons are numbered from 1 to $N_{neuron}$. Suppose that the number of codewords S is limited. And it forms dictionary T with capacity of $N_{dict}$.

$$T = \{S_1 ... S_{N_{dict}}\}$$
$$S_i = (s_{i1} ... s_{iN_{neuron}})^T$$
$$s_{ij} \in \{0, 1\}$$
$$\sum_j s_{ij} = N_{sig}$$

So, on the dendritic surface of any neurons in the cortex, "selected" place can be found for any of the signals. The high density of neurotransmitter at such place will correspond to the emergence of the specific signal. It turns out that with reasonable capacity of the dictionary these places will have a sufficiently selectivity. This mean to identify the corresponding signals well. Take an arbitrary signal $S_i$ from the set T. Assume that all the signals in the dictionary created randomly. Take an arbitrary neuron, one that have place on dendrite where the density of neuromediator on signal $S_i$ is equal to K or higher. Identify error take place when another signal at same place create at least the same density of neuromediator. Let's calculate the probability that other signal from the dictionary will create density of neuromediator not less than K.

So, signal $S_i$ is defined the trap. State of this trap determined by $N_{trap}$ sources. To make the density of the other signal $S_j$ not less than K, it is necessary that at least K connections of neurons, that are active in the signal $S_j$, be in the trap. For a single connection probability of form the trap

$$\frac{N_{trap}}{N_{neuron}}$$

If m connection already form the trap, the probability will be a bit less

$$\frac{N_{trap} - m}{N_{neuron}}$$

Cases where $m \leq K$ are interested. Since K is in this case substantially smaller $N_{trap}$, it can be assumed

$$\frac{N_{trap} - m}{N_{neuron}} \approx \frac{N_{trap}}{N_{neuron}}$$

That is the scheme of Bernoulli trials. The probability of density equal exactly K in the trap by signal $S_j$ is

$$p(K) = C_{N_{trap}}^{K} \left( \frac{N_{trap}}{N_{neuron}} \right)^{K} \left( 1 - \frac{N_{trap}}{N_{neuron}} \right)^{N_{trap} - K}$$

The probability of density to be at least K for signal $S_j$ from the dictionary T determined by the sum of probabilities

$$P(K) = p(K) + p(K+1)...p(N_{sig})$$

The probability that at least one signal from the dictionary than the original, give density K or higher, that is, in other words, the probability of error is

$$P_{error} = 1 - (1 - P(K))^{N_{dict} - 1}$$

For previously used parameters, and the capacity of the dictionary $N_{dict}$ = 10000 calculated values shown in the table:

Table 3. Probability uniqueness violation at synaptic traps at different levels of neurotransmitter density.

| **K** | **$P_{error}$** |
|---|---|
| 3 | 0.00399 |
| 4 | 1.05E-05 |
| 5 | 1.89E-08 |
| 6 | 2.33E-11 |
| 7 | 0.0 |
| … | |

It turns out that when K = 3 traps have a certain selectivity, though not guarantee against mistakes, but at K = 5 they begin quite clearly detect spatial pattern of activity. Remind, that this is not true for arbitrary signals from an infinite set, but it’s true only for set of discrete allowed states of activity, albeit enough large.

All this allows to suggest that the meaning inherent in the brain structure of axonal and dendritic trees is to build on the dendritic surface of each neuron rich of traps. Locations corresponding to all possible combinations of neurotransmitters sources. By density of neurotransmitter in the traps can be detect the spatial distribution signals, composed by simultaneous activity of number nearby neurons.

By the way, this approach can also try to explain the peculiar spatial form of pyramidal neurons that make up the bulk of neurons in the cortex. Structure of apical dendrite structure and branch, structure of the dorsal dendrites, axon orientation - all contribute, on one hand, the coverage of the entire thickness of the cortex in a vertical direction, on other hand, provides minimal intersection between axon and own dendritic tree.

As part of the combinatorial approach possible to explain the purpose of dendritic spines, which is 75% of the synapses (Braitenberg V., Schuz A., 1998). Well, all synapses could be located directly on the dendrite or neuron body. This does not affect the direct work of synapses and their ability to contribute to the appearance of evoked activity. It can be assumed that the purpose of dendritic spines - is the creation of a spatial structure in which different neuronal synapses are "mixed" so that its spillover acquire the ability to influence not only on their surface, but also the surrounding dendrites. The ratio of synapses on spines and directly on the dendrite, spines random height and their configuration - all this fits in well with such assumption.

## The metabotropic receptors

Induced spikes is one created by ionotropic receptors located in the synapse of a neuron. When signals on sensitive activating synapses, and this does not interfere with signals on sensitive inhibitory synapses, the neuron responds by a series of impulses. Synaptic image that describes such signals are called typical stimulus for neuron.

Additionally ionotropic receptors there are metabotropic one. These receptors are not ion channel. That’s why they do not directly involved in membrane polarization or depolarization. They act indirectly by modifying the activity of ion channels, ion transporters and receptor proteins (J. Nicholls, Martin R., Wallace B., P. Fuchs, 2003).

Metabotropic receptors have seven transmembrane domains, an extracellular amino end and an intracellular carboxyl terminus (figure 9). The second and third loops (C2, C3), located in the cytoplasm, as well as adjoining to the membrane portion of the intracellular end have the ability to contact the relevant G-protein.

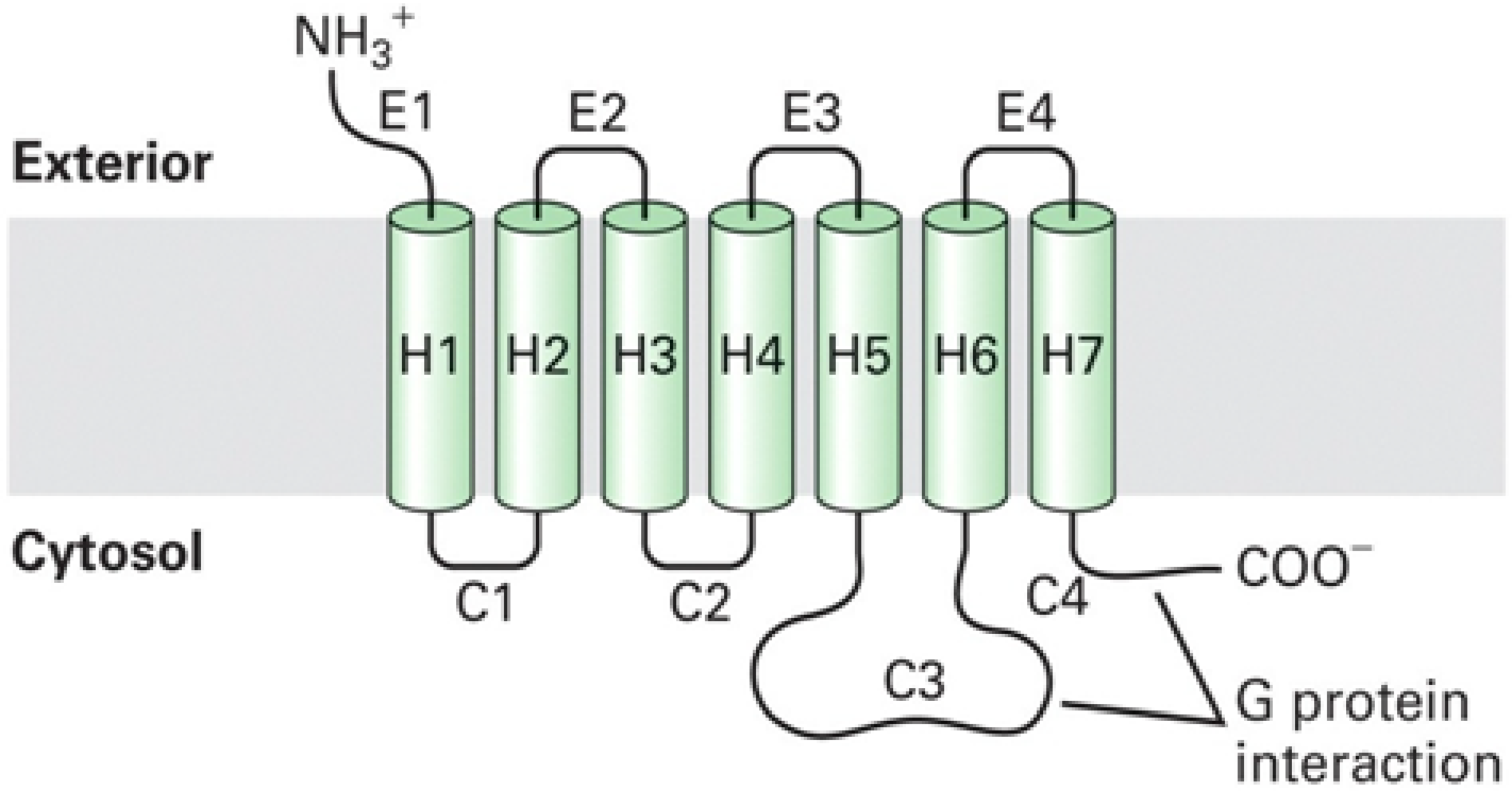

Figure 9. Metabotropic receptor structure

Impact on the neuron's membrane potential metabotropic receptors exert through G-proteins. G-proteins affect ion channels, both directly and with the involvement of second messengers. Engaging secondary intracellular messengers leads to increased efficiency of the receptor in the tens or hundreds of thousands of times. This means that even a small proportion of a substance which is a ligand for the metabotropic receptor may be sufficient to cause a corresponding neuron spike. This mechanism uses by sensory neurons responsive, for example to detect light or odor. Mechanism of signals amplification from metabotropic receptors is typical not only sensory neurons, but the central one too. It is the cause of spontaneous (endogenous) spikes (Nicholls, J., Martin, R., Wallace B., P. Fuchs, 2003).

Typical stimulus for the neuron is input signals pattern that matches synapses sensitivity pattern. Neuron’s evoked activity is reaction on typical stimulus. Evoked activity is only part of the overall brain activity. Another part is so-called background activity, which consists of briefly occasional single endogenous spikes (figure 10).

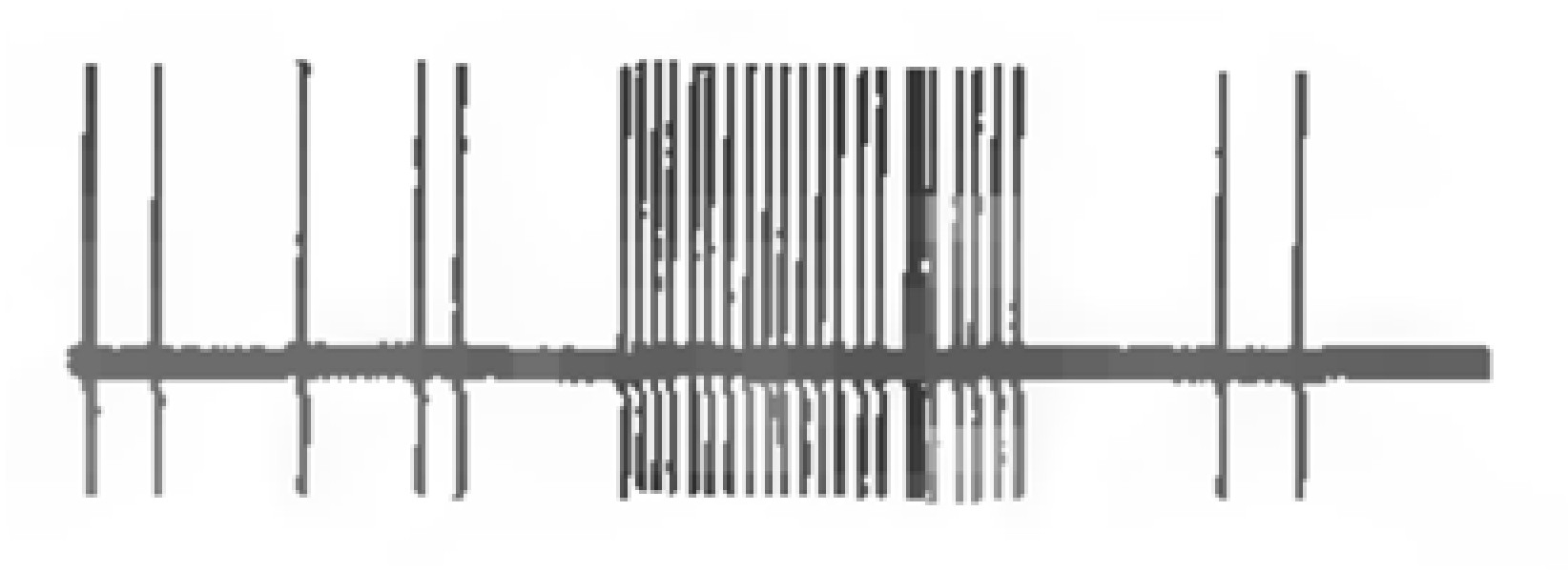

Figure 10. Neuron reaction to the stimulus and the background (spontaneous) activity

Usually evoked activity is a series of impulses, but can be a series of short impulses or even single one. The main difference between evoked and background spike is the level of membrane potential before it. For evoked spike typical pre-spike membrane polarization, but the background one can fire even when membrane hyperpolarized (Figure 11).

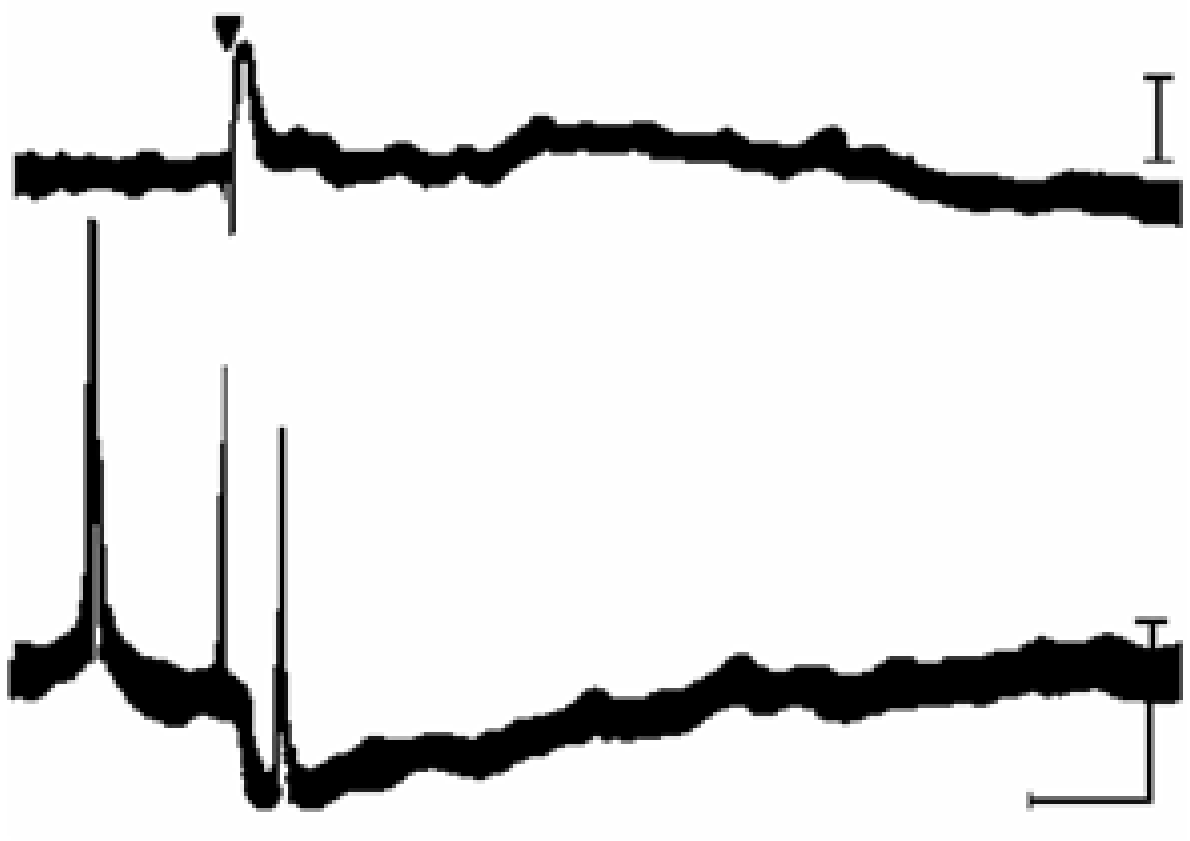

Figure 11. Endogenous spikes may occur independently of membrane depolarization and even opposite - against it hyperpolarization (JP Shnurova, ZM Gvozdikova, 1971)

## State changes of metabotropic receptor clusters

There is reason to believe that the receptors may be quite freely move across the membrane of the neuron (Sheng, M., Nakagawa, T., 2002) (Tovar KR, Westbrook GL, 2002). In the synapse, receptors can "anchor" by contacting postsynaptic densities proteins (Sheng, M., Sala C., 2001). Shown that increasing the sensitivity of synapses, such as their long-term potency, accompanied by migration of additional receptors into the synaptic cleft (Malenka RC, Nicoll RA, 1999).

Outside synapses receptors are moved, including under the influence of neurotransmitters streams formed by spillover. We can assume that it helps receptors grouped in areas of the membrane that are traps for quite regular signals (Radchenko, 2007).

Neighbor receptors can be connected, creating dimers (figure 12). Dimers, in its turn, unite to form clusters of receptors.

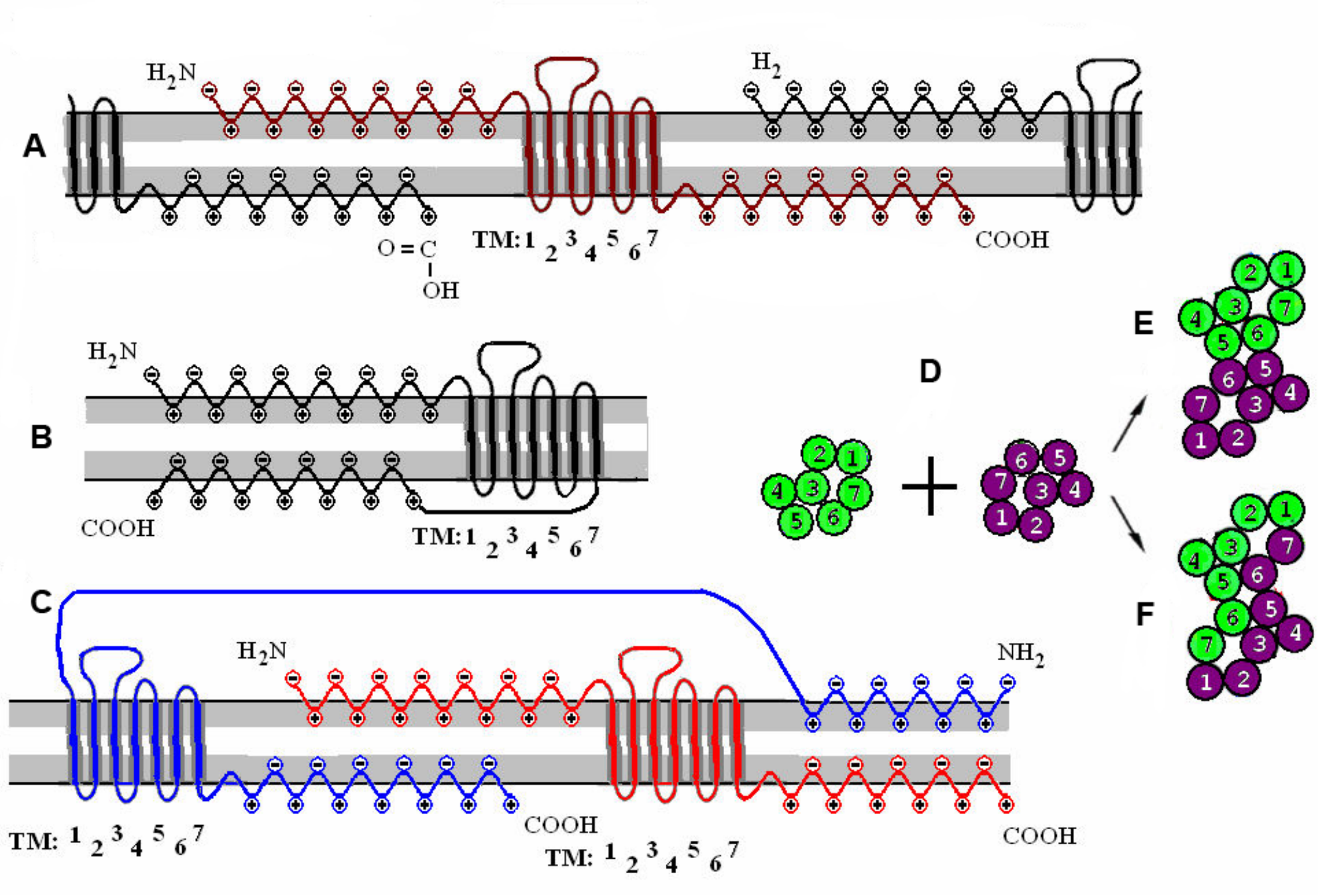

Figure 12. Cluster of receptors. A - single receptor and its interaction with the receptors around. B - monomer receptor molecule. C - receptor dimer. D - union of the two monomers in the pin (E) and the combination (F) dimers. (Radchenko, 2007)

Being in the synapse, receptors located on the postsynaptic densities that prevents appear their metabotropic properties. Outside of synapses corresponding receptors are capable to activate the secondary messanges and thus are potential sources of the endogenous activity.

The difference between the charge of the outer and internal environment creates neuron membrane potential. In addition, on boundary of medium with different dielectric constants arise Donnan potentials. It appear on both the inner and outer boundary of membrane (Coster, 1975).

Donnan potentials within the membrane creates a strong electrostatic field strength of 108 V / m. This field ionizes embedded in the membrane part of split ends of the receptors. As a result, the membrane immersed opposed groups forming transmembrane dipoles (Coster, 1975) (figure 13).

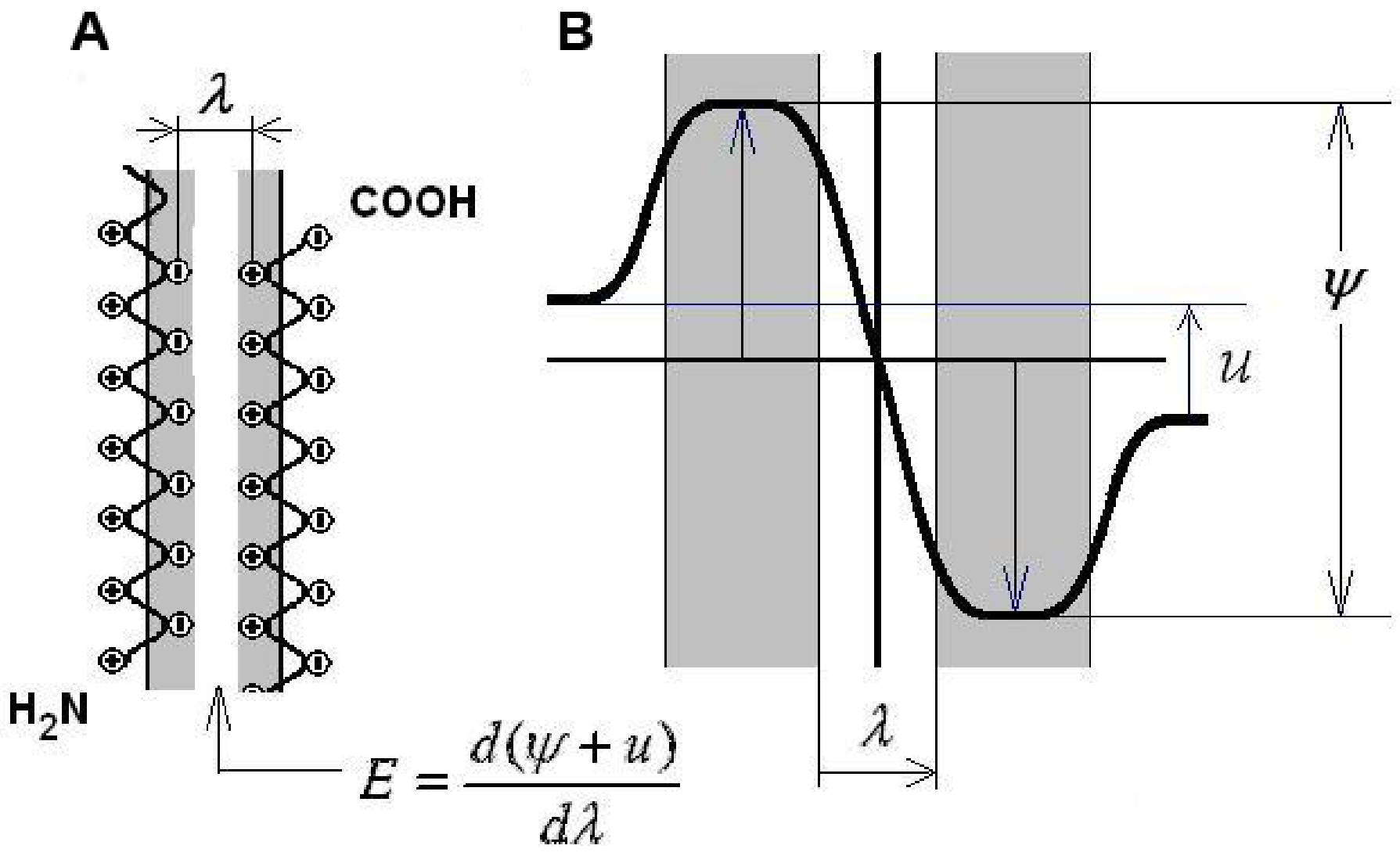

Figure 13. Electrostatic membrane profile. A - transmembrane dipoles. B - the potential distribution, Ψ - the sum of two Donnan potentials, u - membrane potential (Coster, 1975)

Receptors, combined into cluster, like a charged capacitor with plates on elastic suspension (Radchenko, 2007). An important property of such receptors is the ability to change the distance between the transmembrane dipole elements in response to changes in membrane potential or under influence of neurotransmitters.

Hyperpolarization leads to a counter-sinking all receptors ends in the membrane. Depolarization, in contrast, pushes them into the external and internal environment, respectively. Receptor interaction with ligand also changes the distance between ends, pushing them outside and inside the cell.

Immersion external and internal parts of the cluster into membrane weakens the possibility of receptor interaction with ligands and second messengers. Accordingly, the receptors lose sensibility. Ejection out of membrane increases receptors sensibility (Radchenko, 2007).

Interaction of the receptor with the ligand changes its conformation. Conformational changes may persist even after removal of the neurotransmitter. In an altered state of the receptor may be active, that is sensitive to the action of agonists or passive, that is insensitive. It can be assumed that the changed state of the receptor can be reset to its initial state, for example, by strong changes in membrane potential. Also, it can be assumed that an altered state, conversely, can be recorded and preserved by receptor for a long time due to adhesion processes and polymerization.

Figure 14 shows a possible layout of the conformational transitions.

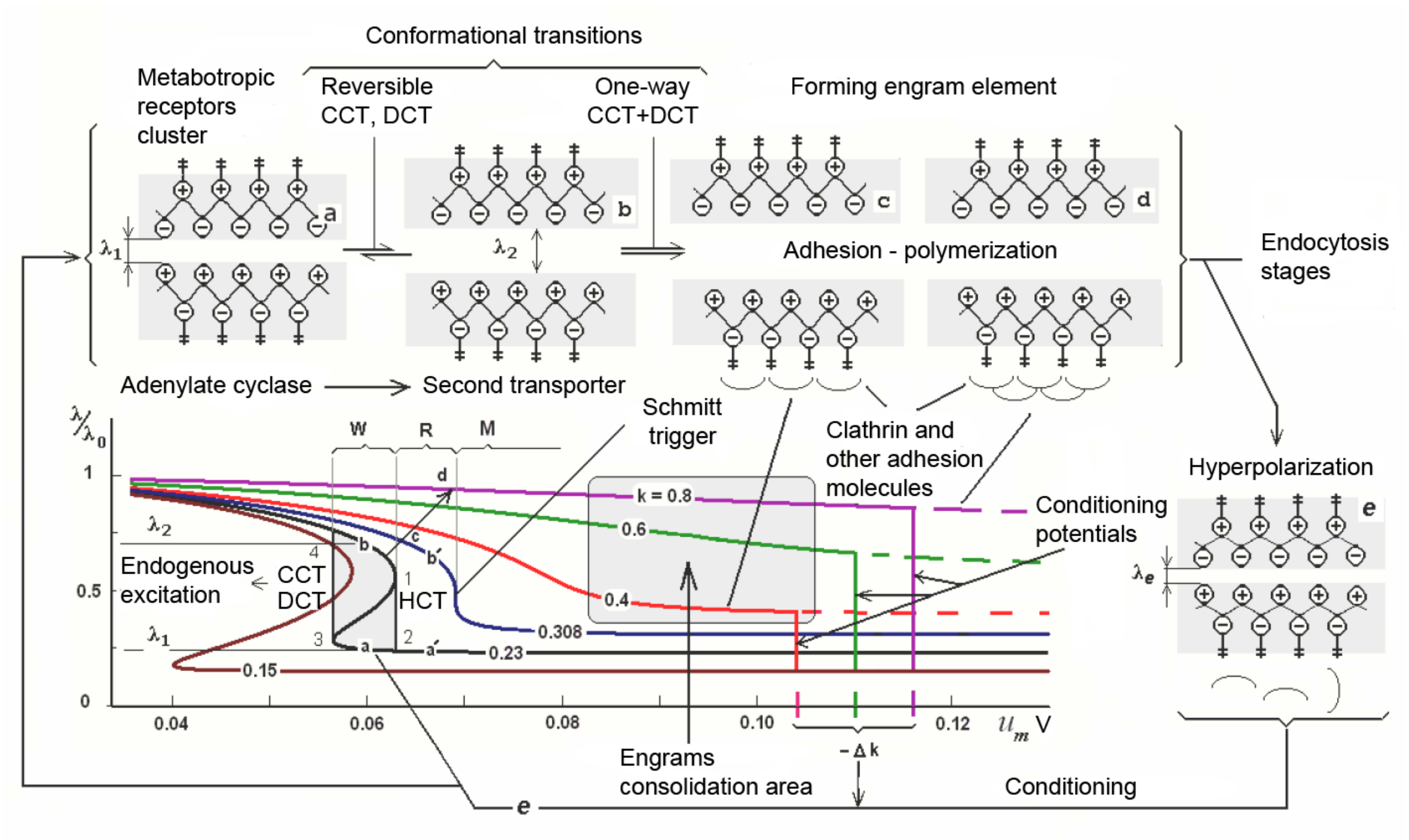

Figure 14. Various state transitions of metabotropic receptor clusters. (Radchenko, 2007)
CCT - chemically induced conformational transition
DCT - depolarization conformational transition
HCT - hyperpolarized conformational transition
λ - the distance between the different parts of the charged ends of receptor

Today, it is difficult to reliably describe all the mechanisms governing changes in the metabotropic receptor clusters. Therefore formulate a number of assumptions:

- Somatic and dendritic membrane of the neuron contain hundreds of thousands of metabotropic receptor clusters. Located in traps formed by various sources of diffusion of neurotransmitters, they are the receivers of these signals;
- To activate the second messenger and start the endogenous spike metabotropic receptor cluster must remain in an active state and thus find themselves in a trap, where there was a significant density of neurotransmitters;
- Apparently, there is a very flexible mechanism regulating state transitions of metabotropic receptor cluster from sensitive to insensitive and vice versa;
- Apparently, there is a mechanism of consolidation, which determines the long-term sensitivity or insensitivity state of cluster. This mechanism controls whether these changes are short, medium or long term.

Assumptions becomes clear if we recall the combinatorial structure of the dendrite. In result, neuron is able to remember the picture of impulse activity of adjacent neurons by changing the sensitivity state of metabotropic receptor cluster on it.

When we are dealing with a limited set of possible activity patterns, traps, which have significant density of neurotransmitter reliably identify the corresponding signals. This means that if metabotropic receptor cluster, located in the corresponding trap, change to sensitive state and set high threshold of activation. Then neuron will respond to the appearance of the signal corresponding to this trap by endogenous spike. Conversely, if the cluster state change to sustainable insensitive mode, the neuron will be blocked and it do not response on this particular signal.

This signals memorization do not directly related to the sensitivity of the synapses. It does not depend on typical stimulus for neuron. One that causes it evoked activity. However, relatively synaptic plasticity known that when the synapse changing sensitivity reshape the structure of postsynaptic membrane of the receptor is reshape, and also the amount of neurotransmitter released by axon become different. That may disrupt traps associated with this synapse. But it is possible that outer side axon emission of neurotransmitters and surrounding glial cells compensate these changes.

Signals memorization at metabotropic receptor cluster has some interesting features. Suppose that signal $S_i$ from the dictionary T have memorized. Assume that in the cortex emergence of a new information state, which is a combination of a small number of same dictionary signals. The combination in this case is binary operation "OR" of binary vectors. For such a signal, neuron still finds signal $S_i$, if it is contained in this combination and it will react by spontaneous spike.

Such an information construction close its spirit to the Bloom filter (Bloom, 1970) (figure 15). In Bloom filter all elements of the initial set are mapped by binary hash functions. Then, for a certain subset by "OR" operation form single binary vector. Using this vector, it is possible with high reliability detect whether given element belongs to this subset, without iteration over all elements of set. This detection can be done only by check out the bits corresponding to hash functions of interest element.

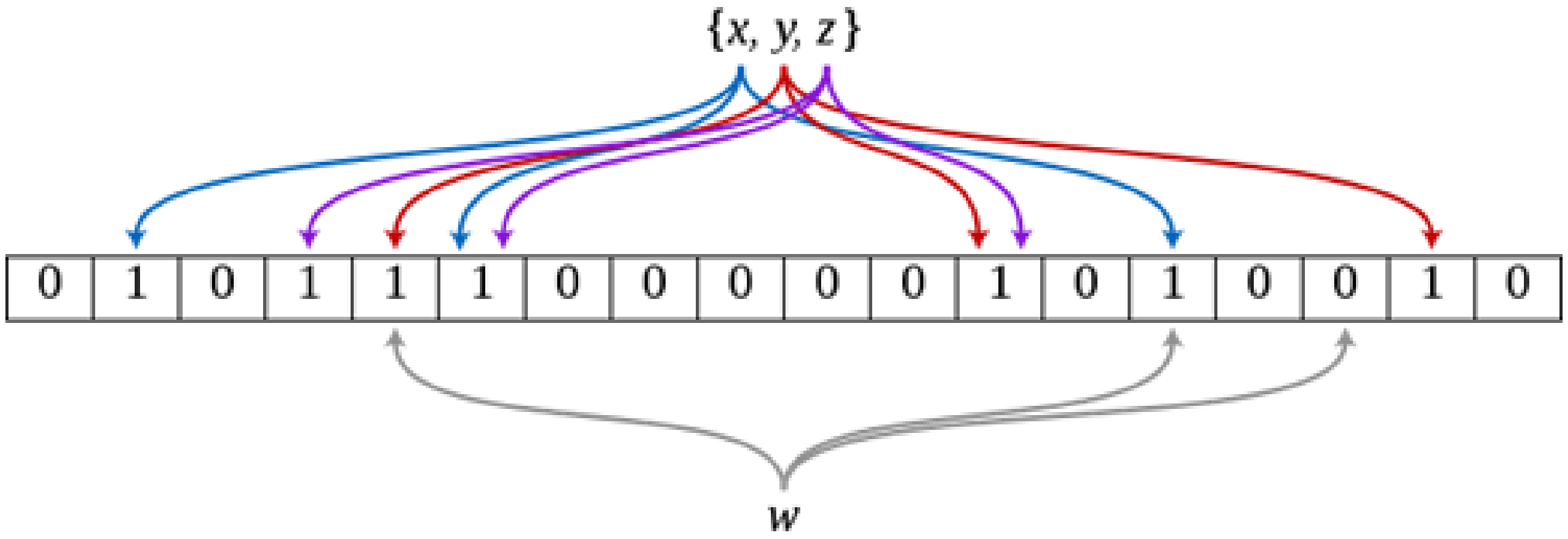

Figure 15. Bloom filter. Element w does not belong to a given set {x, y, z}

Something similar happens in the trap containing the changed state cluster. To detect particular signal no needs to check all neurons around, but rather ensure activity some set of sources.

Once again, note that this is not true for all types of signals. Only one that are made up of a relatively small number of pulses and form a finite set of states.

## Wave activity

Work of brain's neurons is characterized by a certain rhythm. Total changes in membrane potential of individual neurons in cortical areas form a sufficiently clear rhythms, which are recorded on an electroencephalogram. It is believed that this should be impulse activity as well as slower oscillations of the membrane potential (Pribram, 1971).

Methods of optical observation of activity of the cortex allow to trace the spatiotemporal structure of these oscillations. In experimental animals expose portion of the cortex and injected a special dye that is sensitive to changes in the electric potential. Under the influence of fluctuations of the total membrane potential of neurons this dye changes its spectral properties, which may be recorded, e.g. by a diode array which act as high-speed video camera. Optical methods can not look inside the cortex and track the activity of individual neurons, but they enable a general idea of the distribution of wave processes on its surface (Michael T. Lippert, Kentaroh Takagaki, Weifeng Xu, Xiaoying Huang, Jian-Young Wu, 2007).

Optical observations show that brain rhythms correspond to waves arising in certain sources and extending further inside cortical areas. The figure below (Figure 16) shows the phase of wave propagation at 5 millimeter part of the cortex of the rat brain.

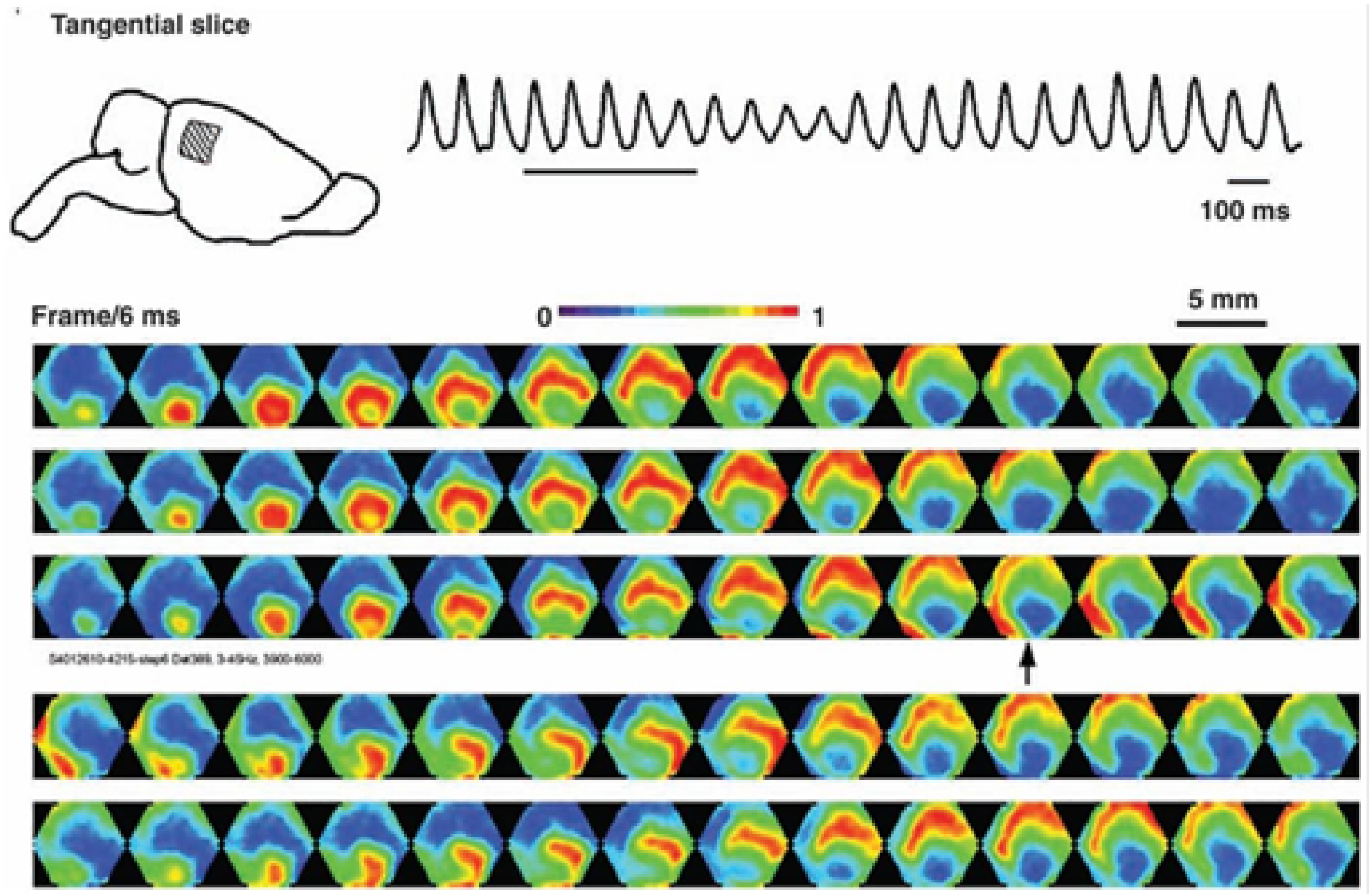

Figure 16. Painting of the wave activity in the area of the cerebral cortex of the rat. Potential shows gradient from blue to red. 14 frames at intervals of 6 milliseconds cover one cycle of the wave (84 milliseconds - 12 Hz) (Michael T. Lippert, Kenta)

It is shown that waves can thicken, reaching the boundary of the cortex, can be reflected from the other zones and create the oncoming wave, can propagate double spirals and create vortices (W.-F. Xu, X.-Y. Huang, K. Takagaki, J .-Y. Wu, 2007).

State of single neuron described Hodgkin-Huxley model (Hodgkin, 1952). This model defines the process of auto-wave in the active medium and explains the ability of neurons to generate rhythmic impulses.

In the simulation of neural networks using oscillating neurons can achieve different group effects. For example, using oscillators with different natural frequency due to the effect of adjusting the phases possible to demonstrate the occurrence of regular waves synchronization (Kuramoto, 1984).

Modeling neural groups pulsations and the propagation of the wave front in the cortex are designed to discover the mechanisms of transfer and processing of information. It is believed that the information can be encoded by frequency and phase characteristics of neural impulses. At the same time as the main mechanism of signal propagation considered induced neuronal excitation by direct synaptic transmission. Previously made assumptions about the combinatorial structure of the dendrite and the possibility of metabotropic receptor cluster to record its state allows a different understanding of the mechanism and general sense of wave processes in the brain.

To illustrate the assumptions about the work of real brain lets use a highly simplified neural network model. At the same time, try to describe clearly the most basic principles. To do this, not needs to reproduce all the supporting mechanisms exactly as it is peculiar to the real brain.

Rectangular portion of the cortex will be simulated. For simplicity of visualization abandon volume. Make use a flat neural network in which the neurons place at node of regular lattice (Figure 17).

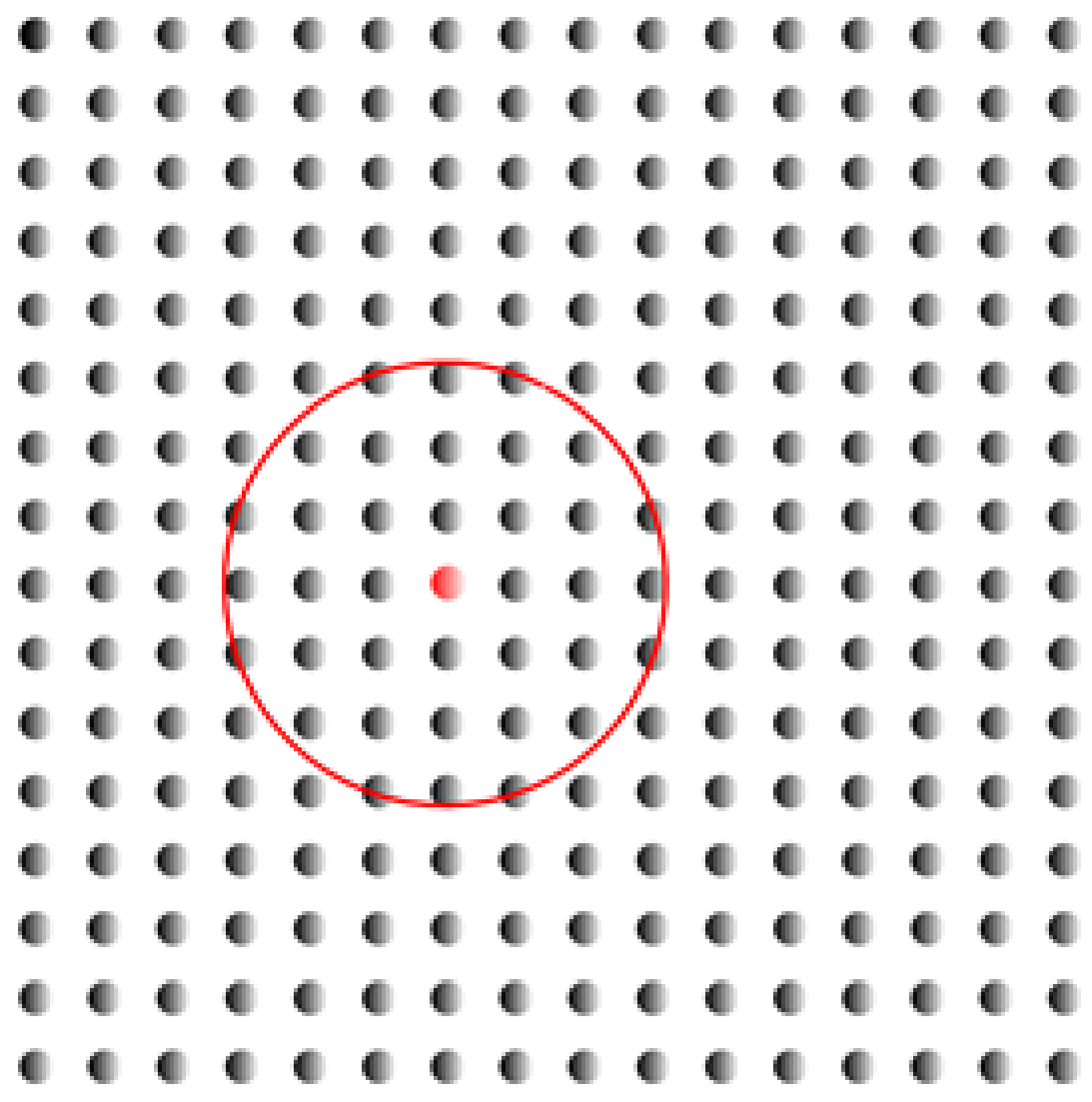

Figure 17. Flat neural network. Red circle is tracking field of selected neuron

Define for all neurons tracking field. Shall assume that each neuron is connected to all the neurons that get into site of radius $R_{obs}$, outside of this radius neuron does not have connection.

Enumerate cortical neurons from 1 to N. The coordinates of the neuron on the net determined by the pair ($x_i$, $y_i$). Tracking area for the neuron with an index can be written as a set of indices that make up the region of neurons.

$$O_i = \{j \mid (x_i - x_j)^2 + (y_i - y_j)^2 < R_{obs}\}$$

The state of cortical activity in discrete time will be simulated. At any given moment the state of neurons can be described by binary vector S.

$$S = (s_1 ... s_N)^T$$

For a single neuron information available to it, the picture looks like the set of pairs (neuron, state)

$$\{(j, s_j) \mid j \in O_i\}$$

Let's simplify mechanism of the neurons synapses work, just assume that due to external causes neurons can come into a state of induced activity. Induced activity of neuron means that its state change from 0 to 1 and persists as long as this evoked activity exist. Evoked activity determines what information arrives to cortex.

Suppose that somehow on cortex arose compact pattern of evoked activity (Figure 18). Compactness means that all active neurons inside the region of radius $R_{obs}$.

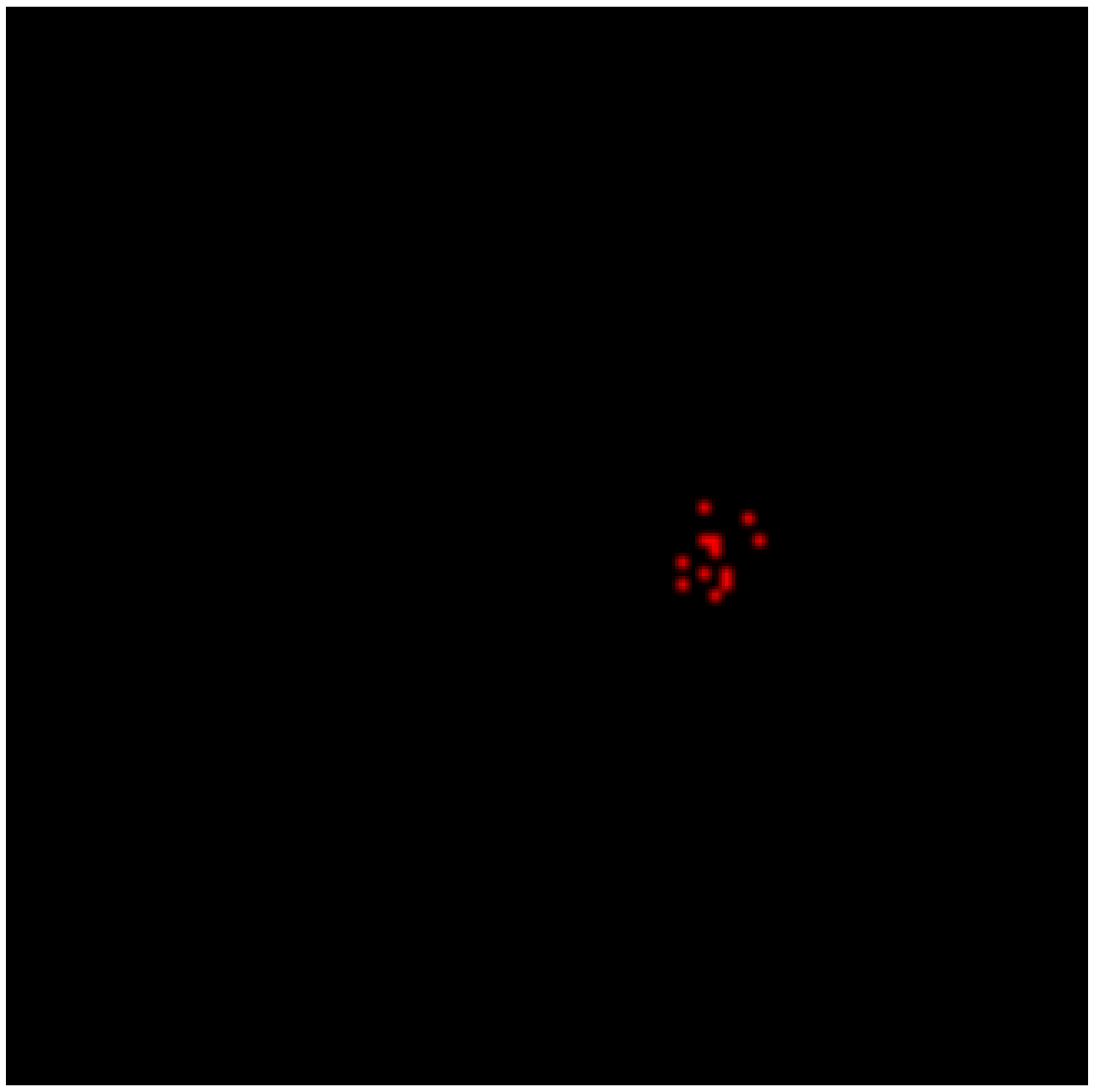

Figure 18. Evoked activity pattern. Showing only active neurons. Neurons are represented without gap. Each pixel corresponds to one neuron.

Now count the number of active neurons in the tracking field each neurons in the cortex (Figure 19).

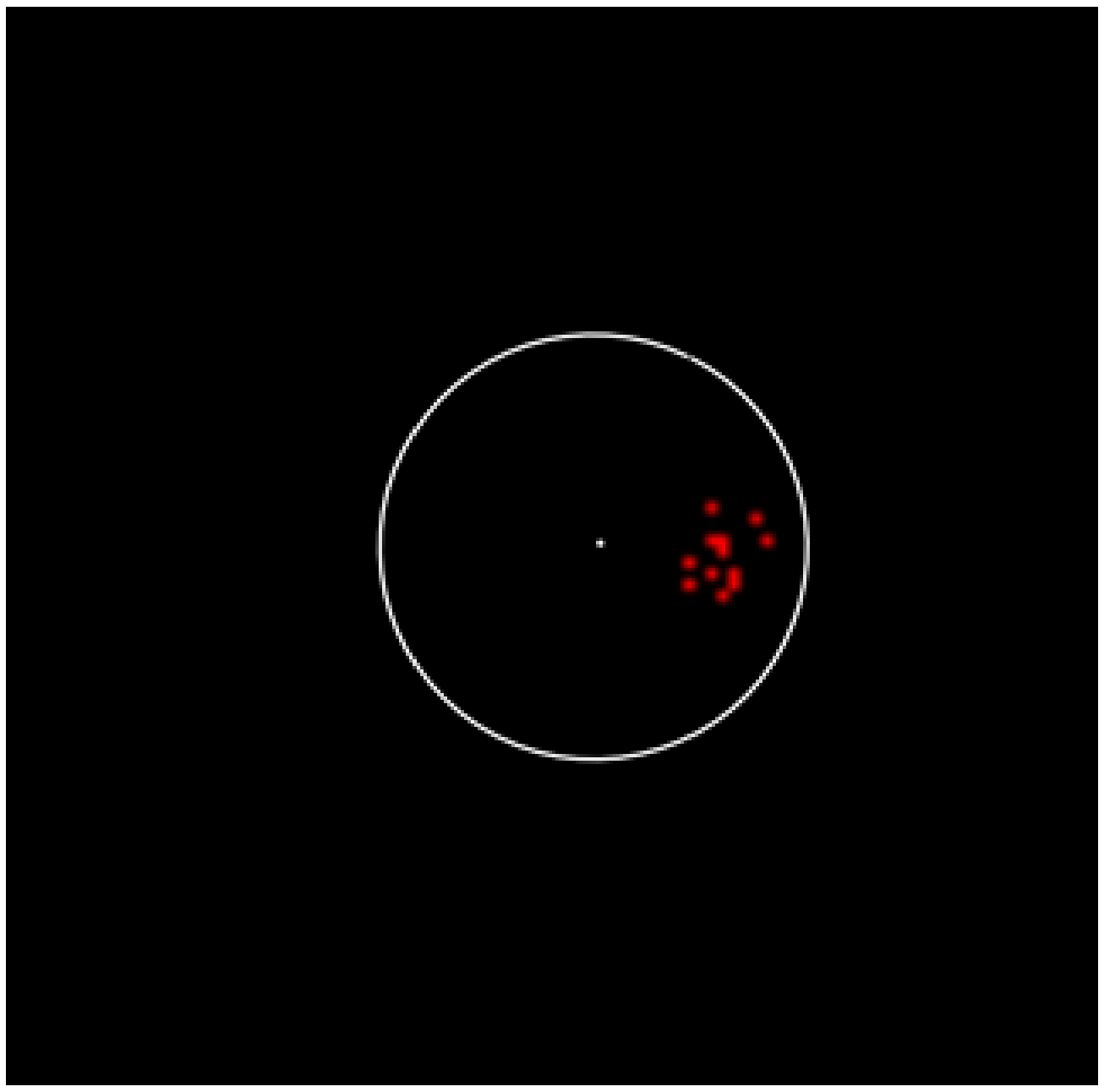

Figure 19. Tracking field for one of the neurons and active neurons inside it For each neuron in quiet state, with number of active neurons in the tracking field exceeds a certain threshold, perform the following procedures. Specify some small probability $p_{in}$. Making the neuron randomly with probability $p_{in}$ produce a single, one cycle, spike. Let’s call this spike endogenous. Accordingly, with probability 1 - $p_{in}$ neuron remain inactive. No matter, gave neuron spike or not, remember which neurons were active in the tracking field of selected neuron and how it reacted (i.e. it gave spike or remained calm). Let M be a set of remembered indexes.

$$M = \{j \mid s_j = 1 \wedge j \in O_i\}$$

Denoted by $M^+$ set of neurons react by spike, and through $M^-$ - remained calm.

Assume that each neuron can remember a large number of M. Denote the collection of $M^+$ as $\mathbf{M^+}$ and the collection of $M^-$ - $\mathbf{M^-}$.

$$M^+ = \left\{M_1^+ ... M_K^+\right\}$$

$$M^- = \left\{M_1^- ... M_L^-\right\}$$

Collection of sets $M^+$ and $M^-$ forms the internal memory of the neuron. Note, that this memory is quite different from synaptic weights memory. Synaptic weights allow to remember picture of all neurons activity or passivity in tracking field. By one set of synaptic weights, the neuron is able to remember the single typical stimulus. The internal memory allows you to record many pictures, but this is done not by complete "print" of the entire image in the tracking field, but by recording the trace of important (active) neurons only.

Back to experiment. As a result of that procedure around the pattern of evoked activity creating a randomly generated one (Figure 20, Figure 21).

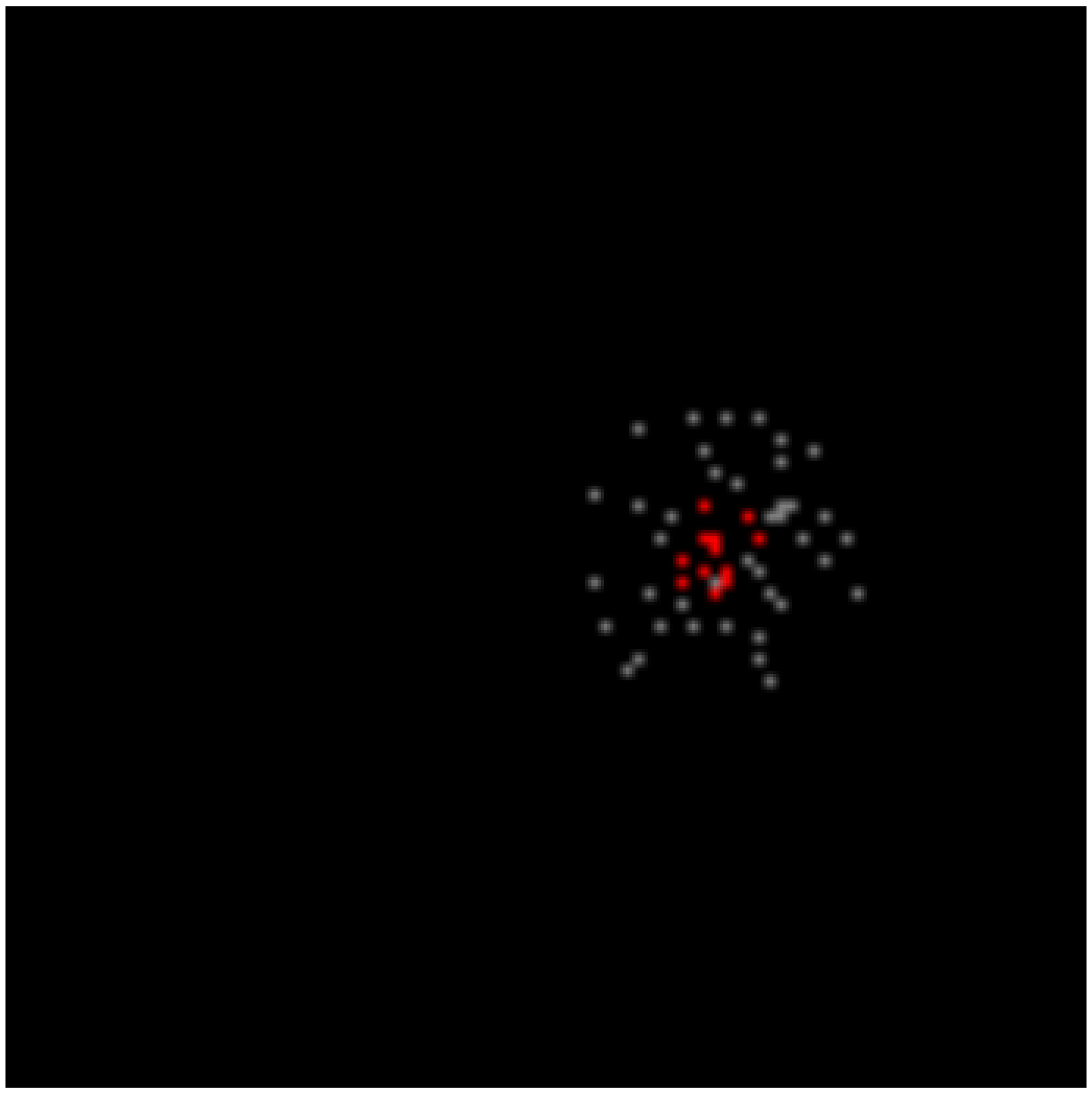

Figure 20. First step of simulation. Endogenous activity (gray) and evoked activity (red)

In the second step of modeling neurons that are located around the periphery of endogenous activity, "see" significant activity in its tracking field. For those who have it exceeds the threshold, repeat the activation procedure.

Neurons activated by the previous step, set state to relaxation. Turn off it activity and for a certain time $T_{relax}$ will block it ability to be excited by memorized pattern.

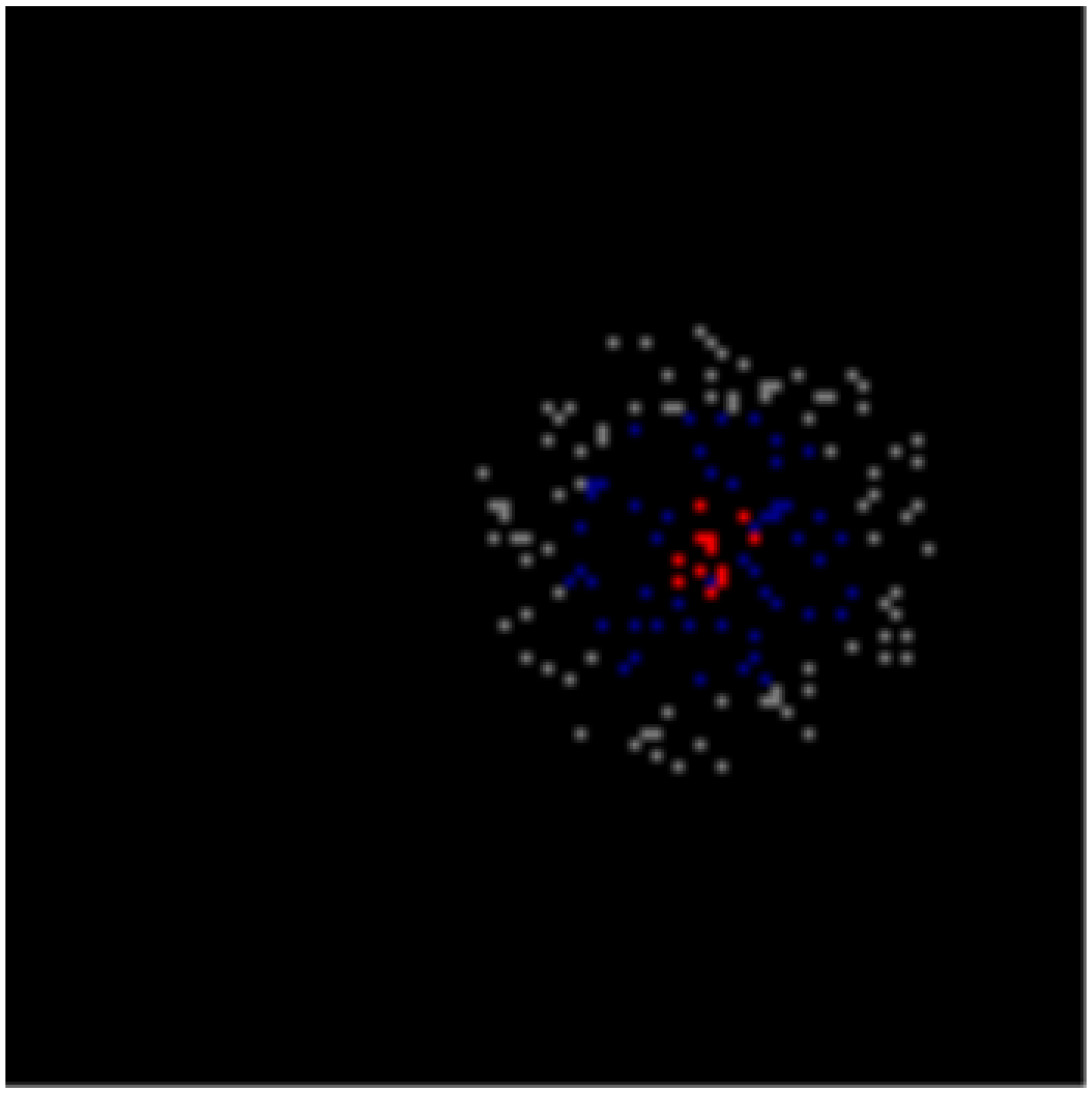

Figure 21. Second step of simulation. Propagation of the wave front. Blue - neurons in relaxation state.

After repeating the steps, spontaneous impulse activity propagating over the cortex will randomly generate certain unique pattern (Figure 22).

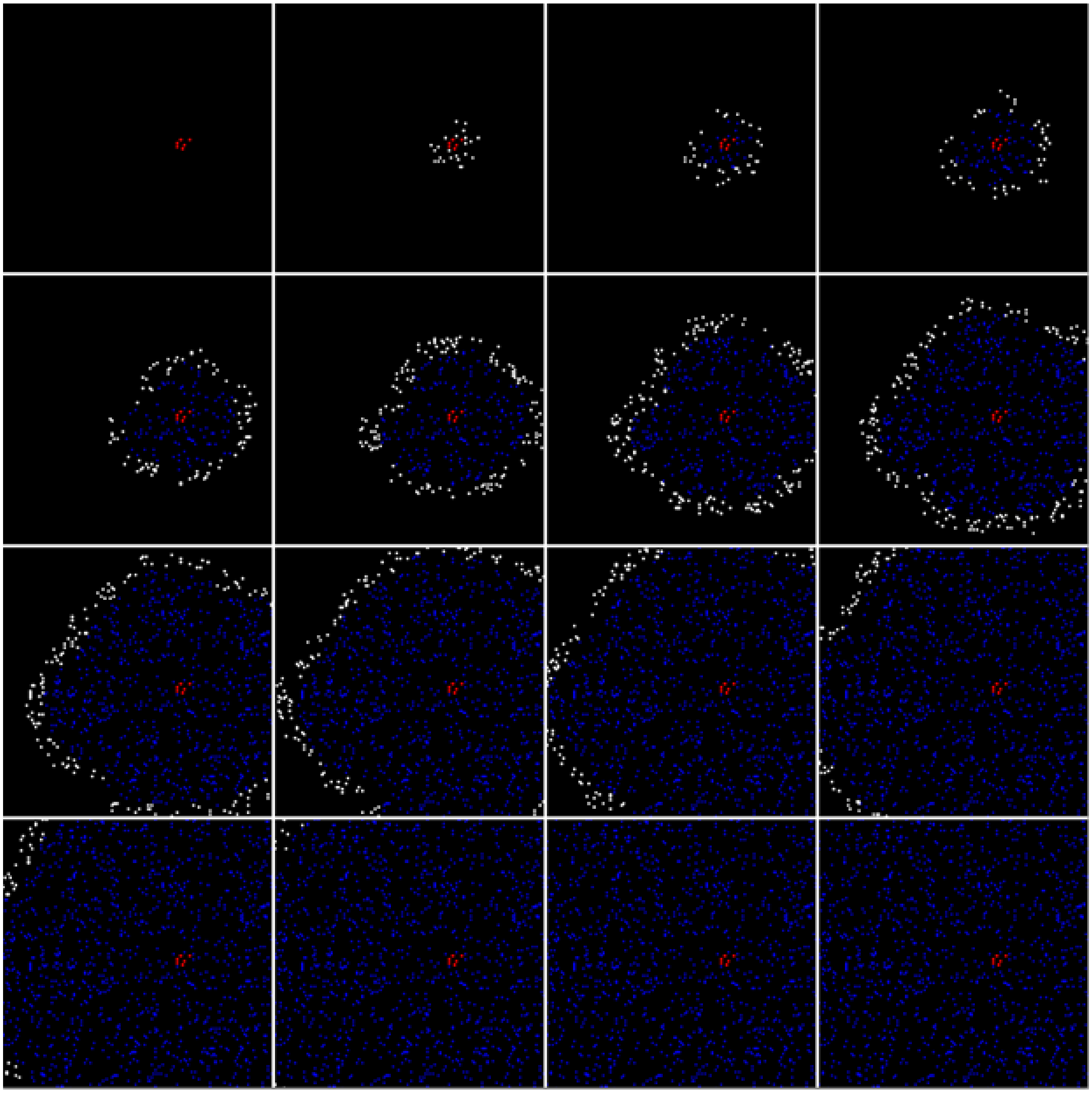

Figure 22. Serie cycles primary propagation of wave pattern of endogenous activity

After the relaxation period passes, new wave will run. But now introduce rule of wave excitation. Each neuron, when detect a high level of activity around, must check does picture of activity match any set of M, remembered in its internal memory $\mathbf{M^+}$ and $\mathbf{M^-}$. Logical function of activity pattern matching S and M element is determined by

$$F_{overlap}(S, M) = \sum_{i \in M} s_i > K_{act}$$

This function takes the true value, when the number of active neurons from the set of M exceeds a certain threshold.

Neuron will generate a single spike when the current pattern of activity matches at least one element of $\mathbf{M^+}$

$$F_{act}(S, M^+) = \vee F_{overlap}(S, M_i^+)$$

If the current pattern gives match with any element of $\mathbf{M^-}$, then exclude neuron from the previously described procedure of single spikes generation. It will not allow neurons initially not participating in the wave pattern, to be added to it in the future.

By that cortex imprinted a unique wave pattern, unambiguously appropriate to the original pattern of evoked activity. Wave, started after relaxation period will repeat the pattern of the first wave (Figure 23).

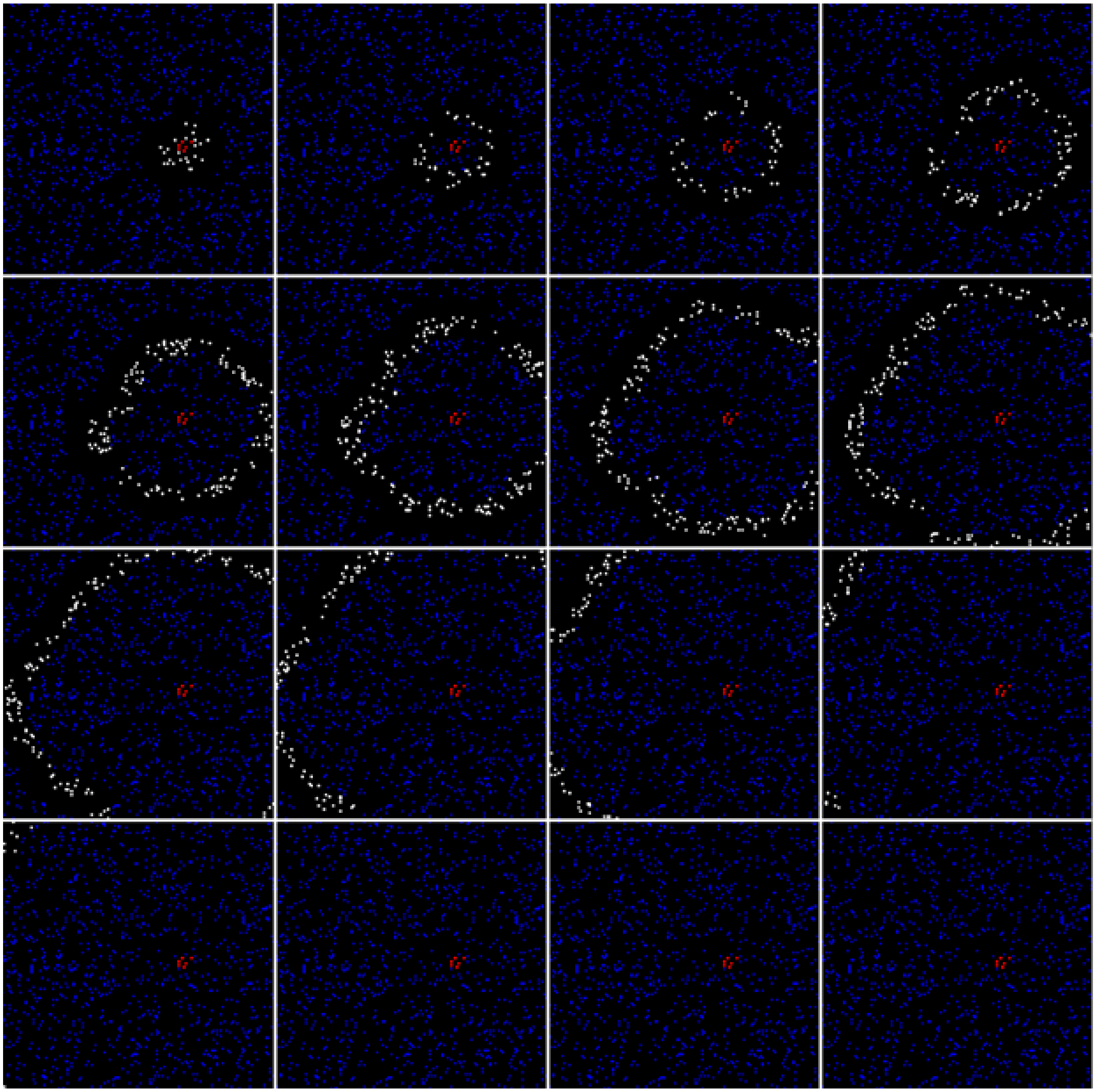

Figure 23. Wave propagation on already “trained” cortex.

Wave will start over and over till evoked activity disappears.

When another compact pattern of evoked activity appear on cortex, it just creates a wave of endogenous spikes propagating away. But, important, that new wave pattern is unique and different from the wave pattern created by previous pattern of evoked activity. Any compact combination of neuronal activity will generate a unique, specific only for this wave pattern.

For each pattern of evoked activity wave will have a unique different from all other wave propagation pattern. And secondly, this pattern is always the same. This means that, after introduce a dictionary T and concepts from the dictionary encode by certain compact patterns of neuronal activity in same, random originally, place of cortex, it possible to determine by passing wavefront which concepts are currently active (Figure 24).

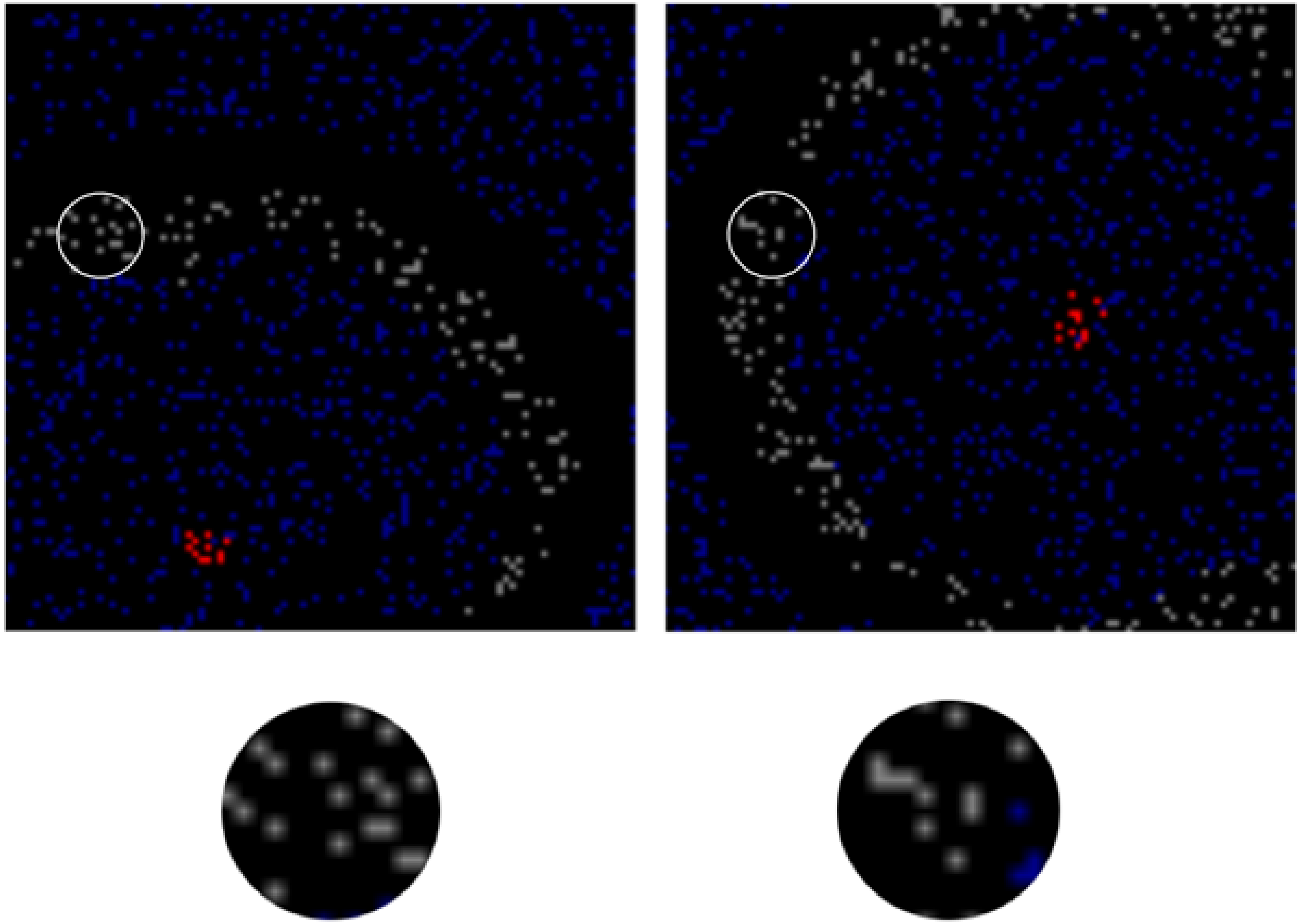

Figure 24. Wave patterns from various initial patterns arising in one and the same place of cortex

Described neural network is well simulated on a computer and shows stable results over a wide range of parameters (Redozubov A., Cortex waves simulation program, 2014).

This computer model can be compared with real cortex. Assuming that the model reproduces the same principles that are peculiar to the brain, then it possible to make the following assumptions.

Any compact pattern of activity is propagated from it a wave patterns of endogenous spikes. The structure of these patterns initially predefined by synaptic connections. Randomly distributed multiple synapses forming traps provide heterogeneous neuronal selectivity towards different surround signals. Some neurons are more "sensitive" to the particular signal than others. This diversity allows any signal to create a corresponding wave pattern on cortex.

For any parameters cortex there is a sharp transition of probabilities of the trap with a given density:

Table 4. Variations of cortex parameters. The probability of finding at least one trap with density K. Highlighted in red level "elite" neurons

| K | $N_{source}$ = 25000<br>$N_{neuron}$ = 650<br>$N_{trap}$ = 15<br>$N_{sig}$ = 10 | $N_{source}$ = 25000<br>$N_{neuron}$ = 400<br>$N_{trap}$ = 15<br>$N_{sig}$ = 10 | $N_{source}$ = 50000<br>$N_{neuron}$ = 650<br>$N_{trap}$ = 15<br>$N_{sig}$ = 10 | $N_{source}$ = 25000<br>$N_{neuron}$ = 650<br>$N_{trap}$ = 15<br>$N_{sig}$ = 15 | $N_{source}$ = 25000<br>$N_{neuron}$ = 650<br>$N_{trap}$ = 20<br>$N_{sig}$ = 10 |
|---|---|---|---|---|---|
| 0 | 0.984 | 0.975 | 0.984 | 0.977 | 0.985 |
| 1 | 1.000 | 1.000 | 1.000 | 1.000 | 1.000 |
| 2 | 1.000 | 1.000 | 1.000 | 1.000 | 1.000 |
| 3 | 0.996 | 1.000 | 1.000 | 1.000 | 0.999 |
| 4 | 0.287 | 0.872 | 0.482 | 0.777 | 0.576 |
| 5 | 0.016 | 0.145 | 0.033 | 0.098 | 0.051 |
| 6 | 0.000 | 0.008 | 0.001 | 0.004 | 0.002 |
| 7 | 0.000 | 0.000 | 0.000 | 0.000 | 0.000 |
| 8 | 0.000 | 0.000 | 0.000 | 0.000 | 0.000 |

At a certain critical value of the the density of neurotransmitter only a small percentage of neurons is "favorites" in relation to a particular signal. Actually, these "favorites" neurons define a unique wave pattern.

If the trigger level of metabotropic receptors of traps $K_{limit}$ close to a level corresponding to a sharp transition probability, it will ensure that small percentage of neurons in the wave. At described simulation this reproduced by specifying the probability of entering the pattern $p_{in}$, corresponding to $p(K_{limit})$.

To keep wave going the percentage of neurons that can be triggered by the threshold density should be adequate to signal density $N_{sig} / N_{neuron}$. In configurations of cortex where $N_{sig} / N_{neuron} < p(K_{limit})$, the wave is damped. When $N_{sig} / N_{neuron}$ a bit bigger then $p(K_{limit})$ wave will propagate, density of signal going to level predefined by parameters of cortex.

If you rely only on the natural randomness of the distribution of neurotransmitters sources inside traps, then such a biological system may be unstable over time. Growth of axons and dendrites, the emergence of new synapses, dying of old neurons - all this will lead to a strong change of wave patterns. But record all the possible combinations of activity over time is entirely do not required. The number of signals dealt with by real cortex although large, but limited. For those signals that the cortex really encountered, it possible to describe the mechanisms that prevent any changes of wave patterns.

Long time wave pattern fixation is possible, for example by making the metabotropic receptors in the traps sensitive on neurotransmitter density exceeds $K_{limit}$, and receptors in traps where this level a bit low $K_{limit}$ (i.e., between $K_{limit}$-1 and $K_{limit}$) make insensitive. If these changes are stably fixed, then also will be fixed and stable wave pattern. Previously described mechanisms responsible for changes in the state of cluster of metabotropic receptors may be just the tools that provide long-term fixation of the desired wave patterns.

Changes in receptors occur certain stages and depend on a number of factors. Depending on the combination of these factors either receptors are reset or take sustainable altered state. The set of such stages determines the consolidation process. One can assume that repeatability of the signal is only essential factors of consolidation.

Under certain configurations and network settings, possible get the effect of a gradual learning cortex. With the gradual training initial wave propagates relatively close and quickly damped. Every subsequent wave propagates a bit further, extending the boundaries. And so on till the wave pattern cover entire surface of the cortex. Such a mechanism is particularly interesting for consolidation process.

In the simulation was reproduced several different algorithms creating a sustainable wave patterns. All of them have shown their efficiency. The algorithm described above - is one of the possible algorithms, the most elegant in terms of the author. However, if nature is really chose this wave model for information processes, there is no doubt that during evolution it honed to perfection, creating a lot of mechanisms that increase its reliability and efficiency. Therefore, probably does not make sense to go into the description of a particular software implementation.

An interesting fact is that if the signal is encoded by quite a small percentage of the activity, it allow simultaneous propagation of several waves to be stable. In the propagation of several waves' fronts can pass through each other without changing pattern (Figure 25).

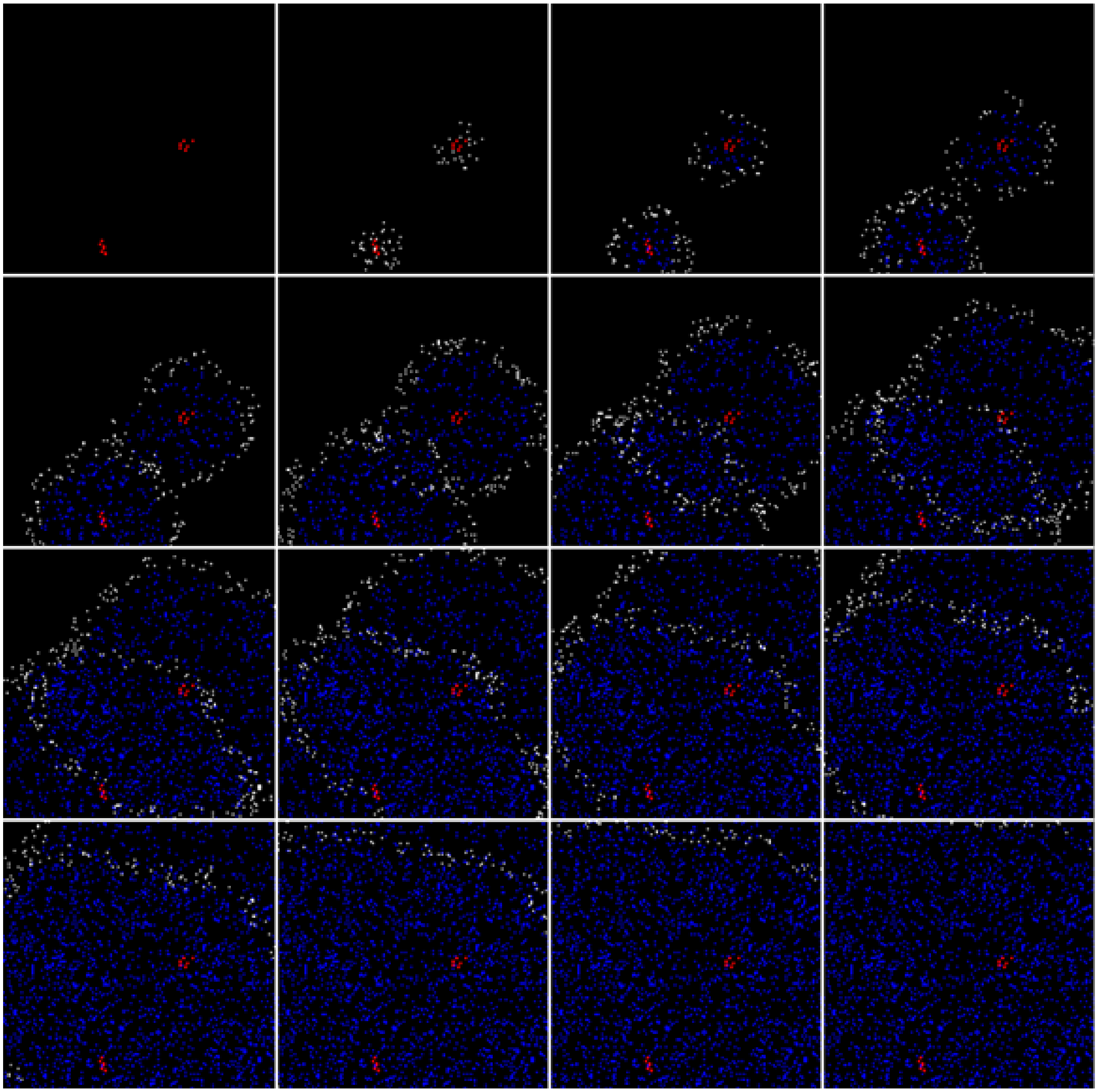

Figure 25. Modeling two sources waves mutual propagation. Cortex was trained on each source separately

It is worth noting that in real cortex relaxation may depend on metabotropic receptor cluster processes rather than neuron one. This means that wavefront propagation do not interfere with subsequent relaxation of the neurons, so that neurons can activated during propagation of other wave.

In the cortex, simultaneously can propagate several waves corresponding to different signals. However, with an increase number of signals in the simulation sooner or later there comes to false positives reaction because of superposition of several wave patterns. False positives reaction avalanche increase the number of neurons involved in this process. At some point, cortex becomes self-excitation regime. This strongly reminiscent an epileptic attack.

It should be noted that the signals are encoded by a cortex, acquire property dualism, which agrees well with the wave-particle duality of elementary particles. Just like particles which exhibit both the properties of corpuscular and wave, the information signal is described in the model - is simultaneously a pattern that starts a wave and a wave that in each phase of the path becomes a pattern which, in turn, continued emits waves.

## Projection system

McCulloch and Pitts focused attention on the fact that white matter fiber bundles connecting the primary visual cortex with other brain regions, are clearly insufficient in scope for the simultaneous transmission of the status of all of the primary cortex neurons (Pitts W., McCulloch WS, 1947). This problem is the narrowness of projecting beams exists not only for the primary visual cortex, but also for all other areas of the brain. Amount of fiber in many projection ways less number of neurons forming a spatial pattern of activity. Communication between zones is clearly not able to pass parallel all the spatially distributed signal.

In conventional multi-layer neural networks, organized like multilayer perceptron (Rosenblatt F., 1962), the state of all neurons forming layer, potentially meaningful. Each subsequent layer for normal operation should receive the status of all neurons of the previous layer.

Something similar can be observed in real projection system of the brain. For example, visual information transmitted by the optic nerve and visual radiation, keeps correspondence that formed by ganglion cells of eye. Such signal transmission can be called analog.

Analog principle applies, perhaps, for all sensory systems of the brain. This allows quite constructively compare their operation with neural networks that also use analog ideology.

However, analog information from the sensory sources, on early stages of brain processing translates into another form. Analog descriptions replaced by "semantic" description. Accordingly, the change form of signal coding and transmission principles.

The proposed concept is assumed that the basic exchange of information between the areas of the cortex base on principles other than those that are peculiar to sensory topographic projection. Areas of the brain are interconnected by relatively thin bundles of nerve fibers forming the white matter of the brain. Moreover, these beams have point contact area with both projecting and receiving zones of the cortex.

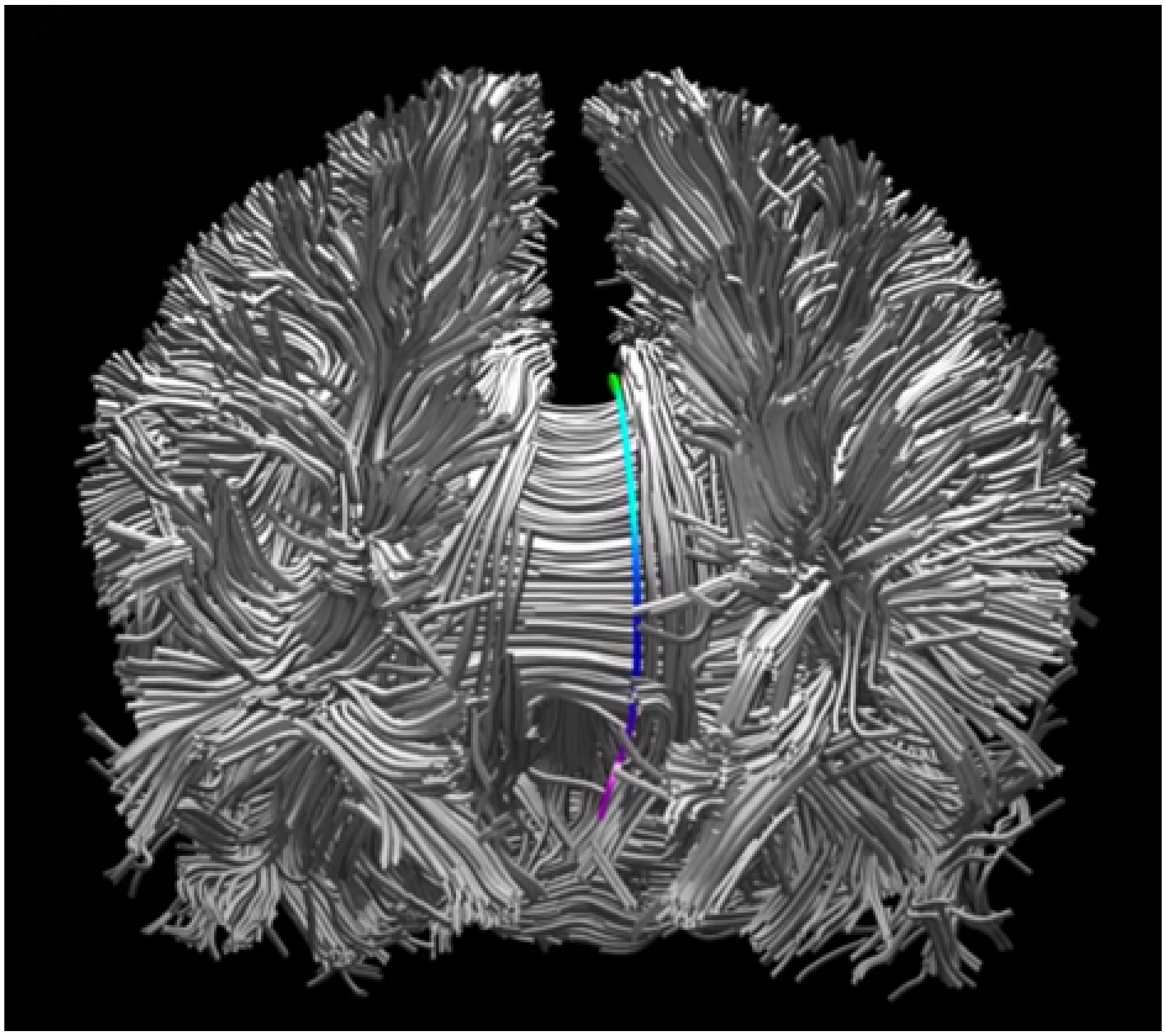

Figure 26. Structure of white matter. Highlighted bundle of fibers connecting two zones (Human Connectome Project, 2014)

White matter fibers can be divided on the projection, associative and commissural. Fiber bundles, going in one direction, form conductive tracts. It can be assumed that the projection tracts carry information in analog form advantageously. Unlike projection, associative and commissural tracts seems, deal exclusively with the transmission wavefronts (semantic) signals.

On figure 27, several images of the same tract of the mouse brain (ALLEN Mouse Brain Connectivity Atlas, 2014). It’s allowing to evaluate connection points size.

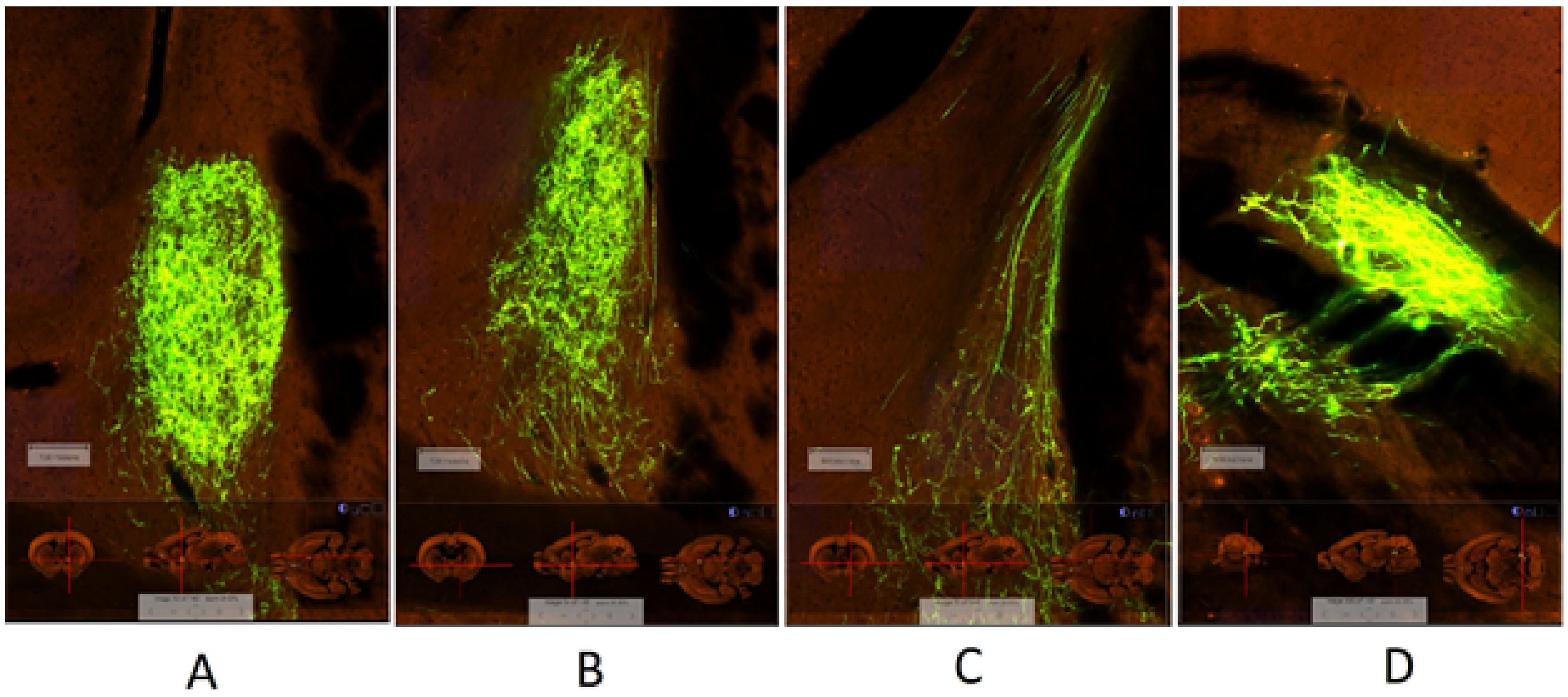

Figure 27. Example of tract. A - the starting point in the cortex, B - transition axons in the white matter, C - fragment of the beam, D - endpoint. Ruler - 100 microns

Let’s show by computer model the possible mechanism of associative and commissural signaling. Take a pair of cortex area and connect them two small, randomly selected areas (Figure 28). Let's make it so that the activity of the upper region is copied to the lower area.

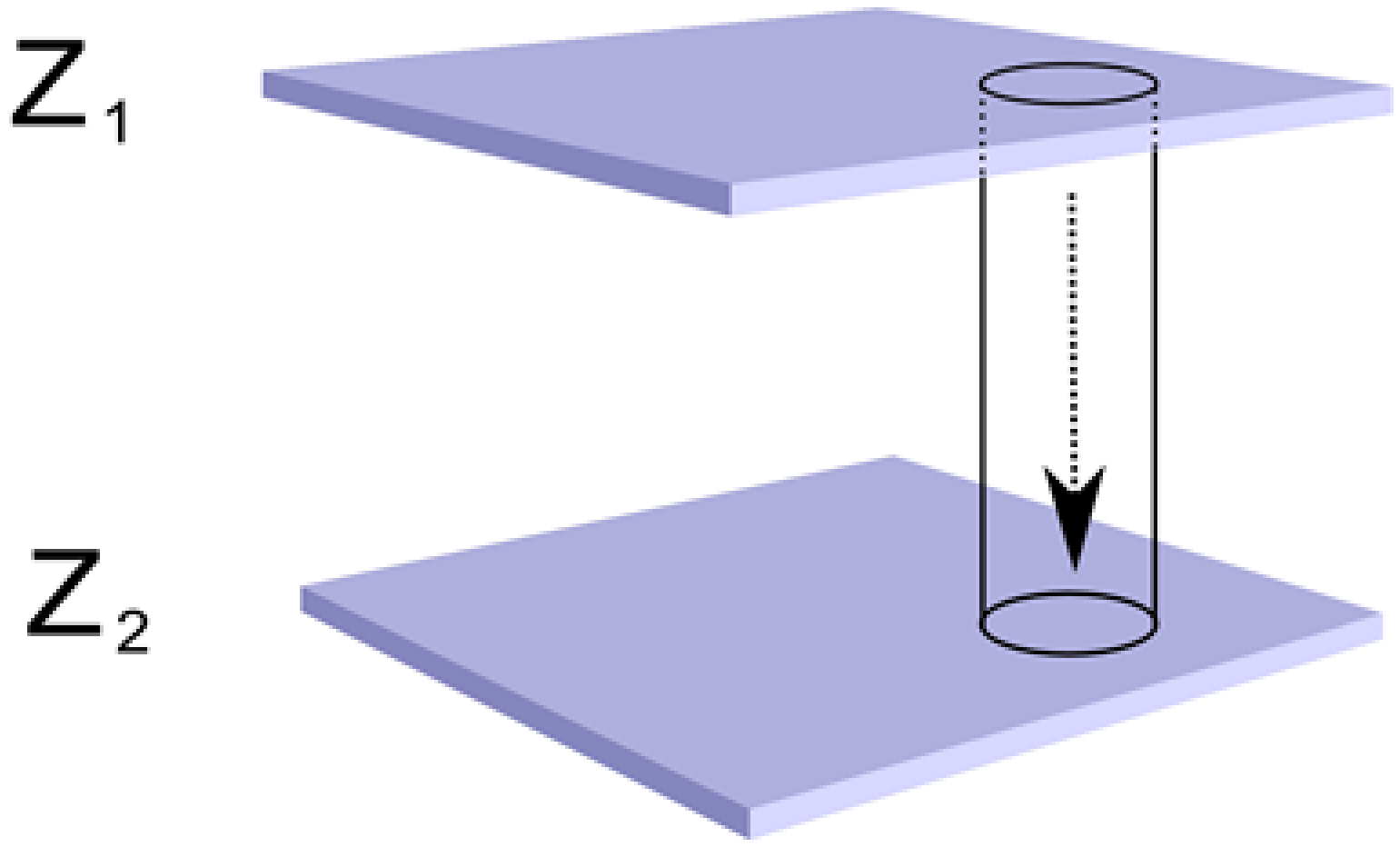

Figure 28. Signal transmission between areas of the cortex (the wavefront tunnel)

In the real cortex part of neurons have axons going beyond the cortex. These efferent axons form tracts. Coming to the cortex that receives the signal, these axons become afferents. Afferent axons form thick collaterals. These branching are similar to conventional axon collaterals (Figure 29). This allows the simulation of white matter fibers by synchronously working pair of neurons. Neuron in the receiving cortex simply duplicates corresponding neuron in the transmitting cortex.

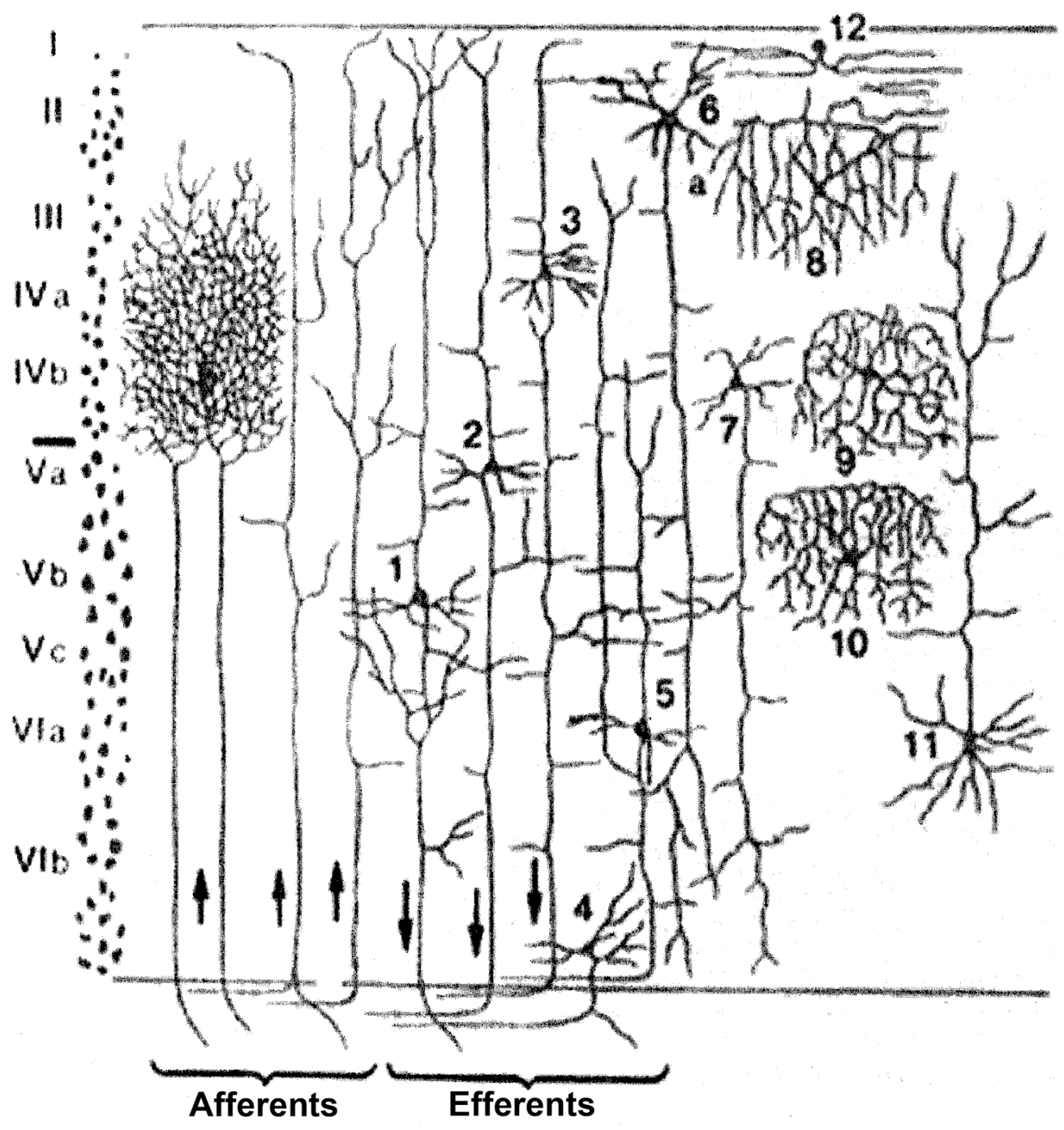

Figure 29. Afferent and efferent fibers of the cortex

Thus, connection between two zones of cortex transfer fragment of activity obtained from one zone to fragment of another zone. Thus especially note that such transformation do not require to be topographical projection. Projecting fibers can be arbitrarily mixed. In addition, some of neurons can be skipped, making the projection depleted. Let's call such a construction wavefront tunnel.

Such a connection can not transfer the whole picture of transmission zone activity. However, such a tunnel is capable of transferring from one cortex zone to another wavefront activity, retaining it rhythm and also the uniqueness of the wavefront pattern.

Passing a small part of the cortex activity from one zone to another, on receiving the zone generating certain patterns of activity that are no different from those that arise in the ordinary wave propagation. This portion automatically become the source of wave. And, importantly, the wavefront pattern on the receiving zone is uniquely associated with a pattern on the transmission zone. One pattern is a natural extension of the other.

This process is observed in the simulation (Redozubov A., Cortex waves simulation program, 2014). Below are two pictures of comparison over time of wave propagation (Figure 30). Top image - transmission zone, lower - receiving zone. Squares highlight areas connected by wavefront tunnel.

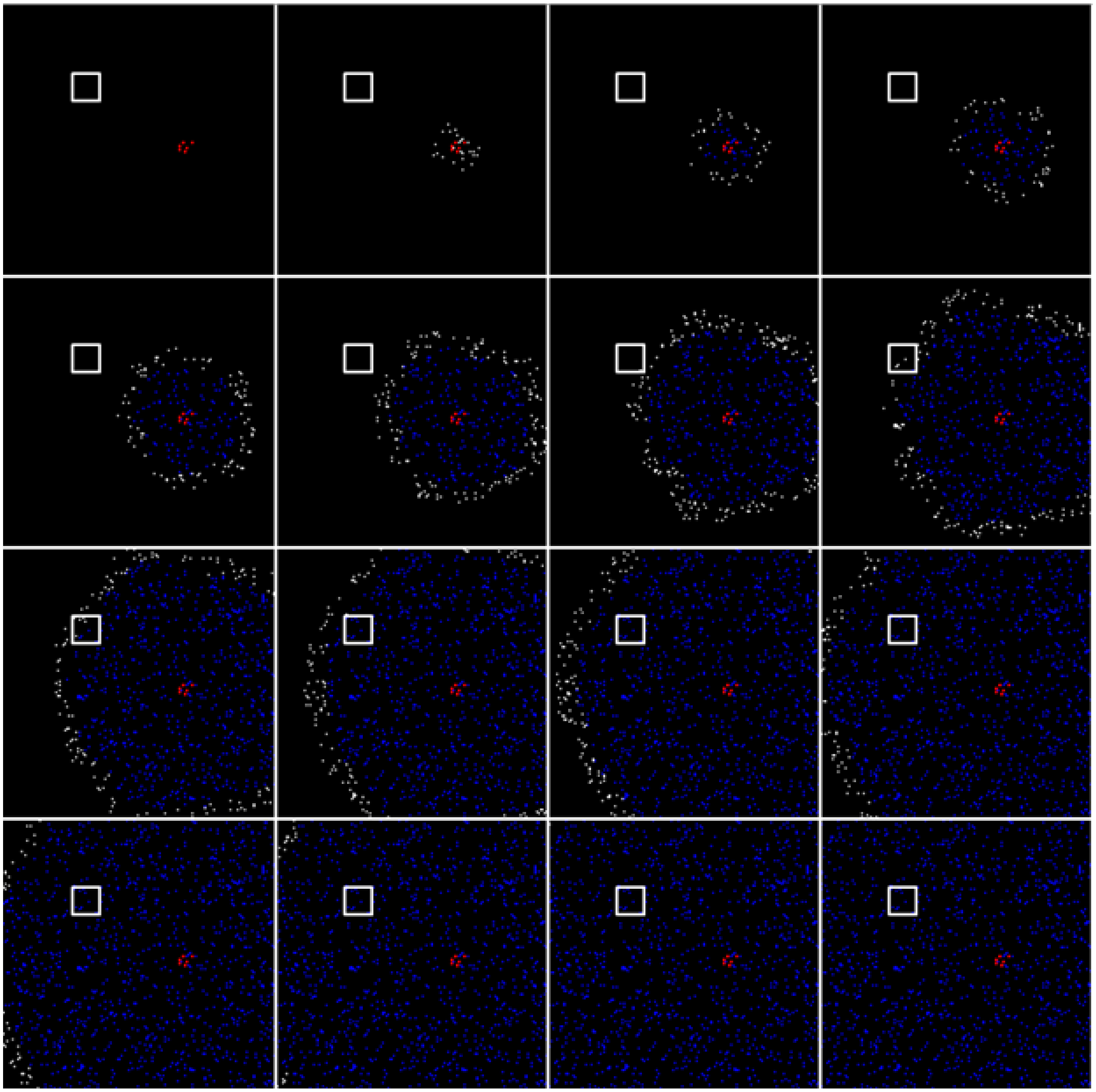

Figure 30a. The transmitting zone activity.

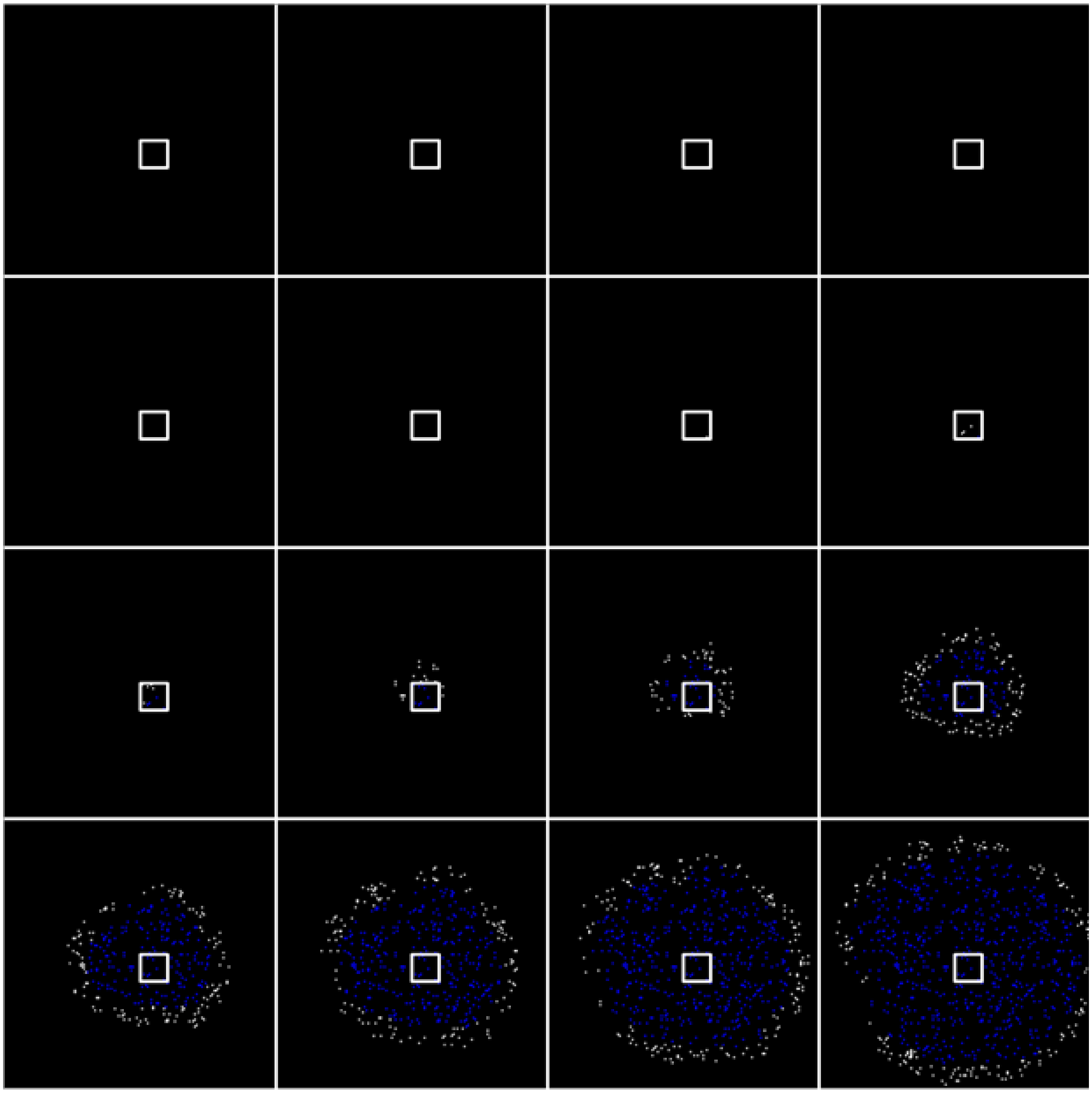

Figure 30b. The receiving zone activity. Images of transmitting and receiving zone are synchronized

It is clearly seen that when the wavefront passes through the tunnel area of transmitting zone, it starts a wave on receiving zone (Figure 31). This wave spreads over the entire surface zone of the cortex. Its pattern is unique and completely determined by pattern of activity that form the projection fiber wavefront tunnel. Since one and the same wavefront at the transmitting zone generates the same output pattern of wavefront tunnel, the wave pattern on the receiving zone can uniquely identify the original signal.

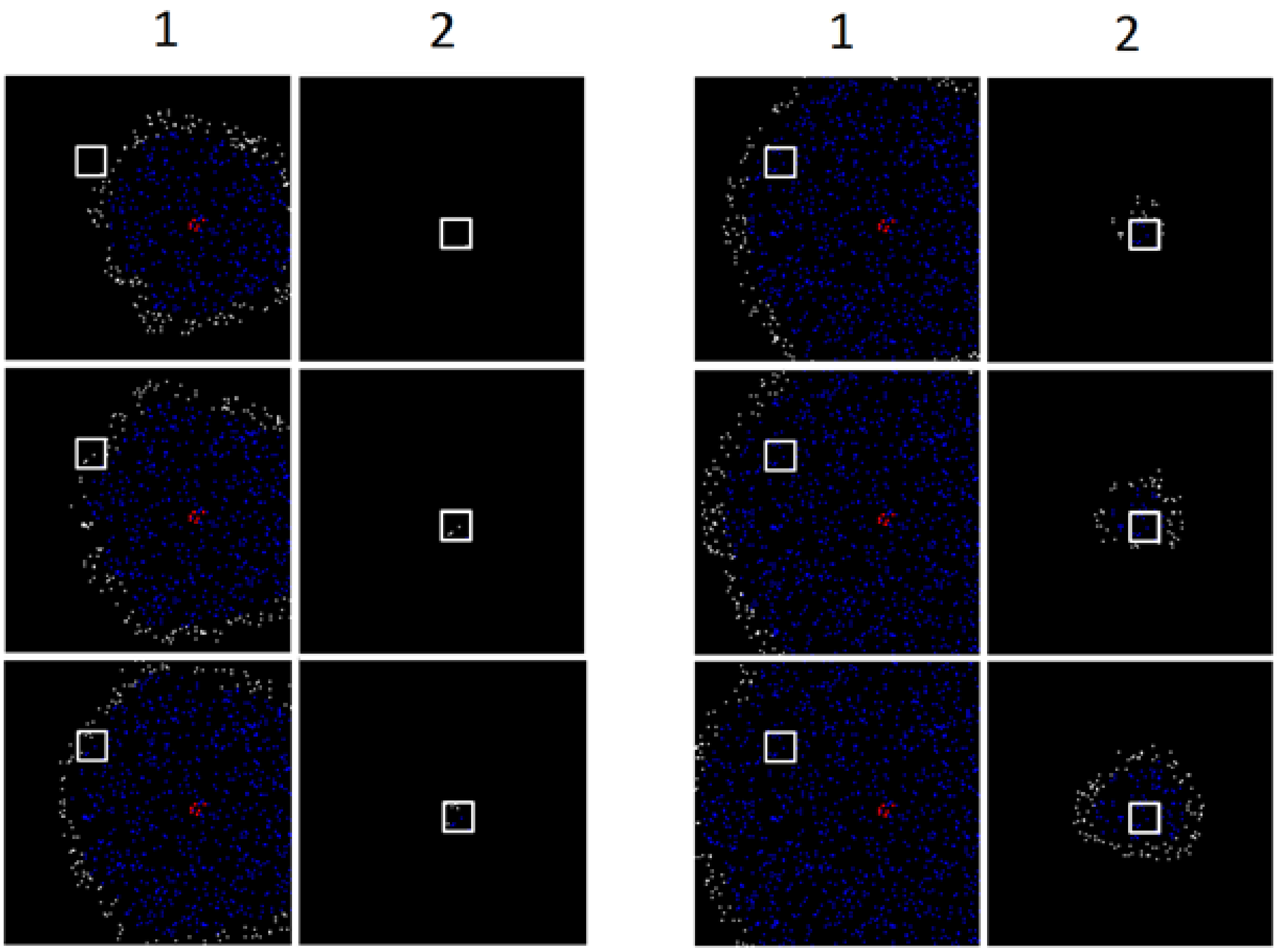

Figure 31. Combining phase of wave propagation. 1 - transmitting zone, 2 - receiving zone

In relation to the real brain described communication mechanism is extremely useful in a number of points:

- to transfer information from cortex this is not needs to transfer pattern with all neurons activity. Enough transmission activity of compact area. The size of it comparable to the size of the tracking field of individual neuron;
- for projection is not critical the point from which place the transmission zone information lifted and the place it comes to receiving zone. Importantly, which zones have a connection between, but it is not important precisely which parts of these zones are connected;
- one cortex can simultaneously have many transmitting and receiving tunnels. Since the linear dimensions of the base tunnel is comparable to the size of the tracking field of cortical neurons (about 200 microns), each cortex may have an extremely high number of connections. Potentially one square millimeter of cortex may contain several tens of wavefront tunnels;
- each individual track can consist of a relatively small number of fibers. Around thousand fibers enough to fully ensure transformation of all information from one cortex to another;
- fibers within a single beam can be arbitrarily intertwine on its route that does not affect the final result.

The general sense of the brain projections, come out from described model, produces the following. Sensory systems of the organism process information flow in analog form. Primary sensory areas converted analog descriptions to semantic one, ie those where information is represented by a finite set of signals and each signal can be associated with a certain sense. In fact, each cortex operates its own set of concepts. Composed of such concepts description is projected to other cortical regions. For secondary sensory, associative and motor areas, input information are semantic descriptions. Set of incoming projections for each zone denine available information to it and, consequently, the system of input concepts. In the process of gaining experience cortex form their own concepts. Descriptions in these concepts form cortex area outputs.

In order to have a complete information about the state of cortex and projected information, neurons do not need to have far-spreading links. For any neuron is sufficient to monitor pattern of activity of immediate around it. By detecting patterns appear around, neuron can adequately “see” what is happening throughout brain. This correlates well with the arguments of the holographic principles of brain (Pribram, 1971). As a hologram, where each fragment thereof contains information about the entire image, each portion of the cortex contains information about everything that describes its cortex, and all that projected on that area.

## Acknowledgment
The author expresses his deep gratitude to Dmitry Shabanov for the constructive discussion, a joint computer modeling and assistance in the preparation and translation of this article.